\pdfoutput=1

\documentclass[11pt]{article}

\usepackage{EMNLP2023}

\usepackage{times}
\usepackage{latexsym}
\usepackage{graphicx}
\usepackage[inline]{enumitem}
\usepackage{float}
\usepackage{subcaption}
\usepackage{booktabs}
\usepackage{amsmath}
\usepackage{tcolorbox}
\usepackage[table]{xcolor}
\usepackage[T1]{fontenc}

\usepackage[utf8]{inputenc}

\usepackage{microtype}

\usepackage{inconsolata}

\usepackage{standalone}
\usepackage{tikz}
\usetikzlibrary{shapes.geometric, arrows.meta, positioning, fit, calc, shadows.blur}

\definecolor{clouddancer}{HTML}{F4F5F6} 
\definecolor{mochamousse}{HTML}{8C5A41} 
\definecolor{surface}{HTML}{FFFFFF}     
\definecolor{borderlight}{HTML}{E2E8F0} 
\definecolor{textdark}{HTML}{0F172A}    
\definecolor{textmuted}{HTML}{64748B}   

\definecolor{accentstate}{HTML}{3B82F6} 
\definecolor{accentpol}{HTML}{8B5CF6}   
\definecolor{accentact}{HTML}{F59E0B}   
\definecolor{accentenv}{HTML}{10B981}   
\definecolor{accentpool}{HTML}{14B8A6}  
\definecolor{accenteval}{HTML}{EC4899}  

\title{Learning to Prompt: Improving Student Engagement with Adaptive LLM-based High-School Tutoring}

\author{Po-Chin Chang\\
  Leiden University \And
  Nicholas Hogan\\
  Futurewhiz\And
  Aske Plaat\\
  Leiden University\And
  Michiel T. van der Meer\\
  Leiden University}

\usepackage{todonotes}

\begin{document}
\maketitle
\begin{abstract}
LLMs can personalize education, although current static-prompt tutoring systems struggle to adapt to diverse academic disciplines. We develop and test a system with subject-aware prompting, based on 14 pedagogical features (e.g., tutor scaffolding, student understanding) extracted from raw transcripts. 
We first train a prompt routing model in a simulation environment, and then deploy it for online adaptation with actual high-school students. The simulation benchmark shows the router outperforming two static baselines ($0.694$ vs. $0.647$ and $0.64$, $p<0.001$). A/B testing ($N=656$ conversations from 359 students) shows sim-to-real transfer where the model switches from analytical to scaffolding learning strategies. Our adaptive prompt selection mechanism improves instructional efficiency in the conversation subset with more turns, maintains pedagogical quality and reduces interactions by around 3 turns ($p=0.007$). While a greedy router achieves a comparable exercise conversion rate with the baseline ($19.1\%$ vs. $19.6\%$), a stochastic router that samples strategies leads to a higher conversion rate ($28.1\%$). Overall, while our framework demonstrates strong performance in simulation, the live deployment highlights the ongoing challenges of translating these gains into statistically significant learning outcomes in real-world settings.

\end{abstract}

\section{Introduction}
Large language models (LLMs) have emerged as powerful tools in education \citep{gupta2025beyond, electronics14234683}, enabling the personalization of the tutoring systems. 
However, most research validates pedagogical efficacy within synthetic environments \citep{nam2025efficient, dinucu-jianu-etal-2025-problem}. While the simulation provides safe exploration, the idealization often overlooks the distributional shift, data sparsity, and complex interactions in real-world education \citep{amodei2016concrete}.

\begin{figure}[tb]
    \centering
    \begin{tikzpicture}[
    >={Stealth[scale=0.8, round]},
    uicard/.style={
        rectangle,
        fill=surface,
        draw=borderlight,
        line width=0.8pt,
        rounded corners=16pt,
        blur shadow={shadow blur steps=20, shadow opacity=5, shadow xshift=0pt, shadow yshift=-4pt},
        align=left,
        inner sep=14pt
    },
    bubbletutor/.style={
        rectangle,
        rounded corners=12pt,
        fill=clouddancer, 
        text=textdark,
        font=\sffamily\scriptsize,
        inner sep=10pt,
        text width=4.5cm,
        align=left
    },
    bubblestudent/.style={
        rectangle,
        rounded corners=12pt,
        fill=mochamousse!10, 
        text=mochamousse!90!black,
        font=\sffamily\scriptsize,
        inner sep=10pt,
        text width=4.5cm,
        align=left
    },
    flowline/.style={
        ->, 
        line width=1.2pt, 
        draw=textmuted!60, 
        rounded corners=12pt
    },
    rewardline/.style={
        ->, 
        line width=1.5pt, 
        draw=mochamousse, 
        rounded corners=12pt
    }
]


\node[uicard, text width=2.8cm] (policy) at (0, 0) {
    {\sffamily\bfseries \textcolor{accentpol}{PROMPT ROUTING MODEL}}\\[6pt]
};

\node[uicard, text width=2.8cm] (state) at (-5.0, 0) {
    {\sffamily\bfseries \textcolor{accentstate}{INPUT ($\mathcal{X}$)}}\\[6pt]
    \textcolor{textdark}{\normalsize\textbf{Context Data}}\\[4pt]
    \textcolor{textmuted}{\scriptsize Subject: Physics\\[0pt] Topic: Floating}
};

\node[uicard, text width=2.5cm] (action) at (5.0, 0) {
    {\sffamily\bfseries \textcolor{accentact}{SELECTED PROMPT ($p$)}}\\[6pt]
    \textcolor{textdark}{\normalsize\textbf{Socratic Method}}\\[4pt]
};

\node[uicard, text width=12cm] (promptpool) at (0, 3.2) {
    {\sffamily\bfseries \textcolor{accentpool}{PROMPT POOL ($\mathcal{P}$)}}\\[6pt]
    \textcolor{textdark}{\normalsize\textbf{20 Pedagogical Prompts}}\\[4pt]
    \textcolor{textmuted}{\scriptsize \textit{``[Role] Act as an AI tutor. Your role is to teach \{Floating and Sinking\} using the Socratic Method...''}}
};

\node[uicard, text width=5.8cm] (environment) at (4, -7) {
    {\sffamily\bfseries \textcolor{accentenv}{CONVERSATION}}\\[6pt]
    \textcolor{textdark}{\normalsize\textbf{Live Interaction}}\\[8pt]
    \begin{tikzpicture}[node distance=0.15cm]
        \node[bubbletutor] (t1) {\textbf{Tutor:} Okay. I see you're studying "Floating and Sinking."};
        \node[bubblestudent, below=of t1, xshift=1cm] (s1) {\textbf{Student:} something stays on top because its density is lower};
        \node[bubbletutor, below=of s1, xshift=-1cm] (t2) {\textbf{Tutor:} Well done! Does ice float on water and does ice float...};
        \node[bubblestudent, below=of t2, xshift=1cm] (s2) {\textbf{Student:} Ice floats on water because the density of ice ...};
        \node[bubbletutor, below=of s2, xshift=-1cm] (t3) {\textbf{Tutor:} Your explanation is exactly right!};
    \end{tikzpicture}
};

\node[uicard, text width=4.5cm] (evaluator) at (-4, -7) {
    {\sffamily\bfseries \textcolor{accenteval}{LLM EVALUATOR}}\\[6pt]
    \textcolor{textdark}{\normalsize\textbf{14 Pedagogical Criteria}}\\[6pt]
    {\scriptsize \sffamily
    \renewcommand{\arraystretch}{1.5}
    \begin{tabular}{@{}p{3.8cm}@{}r@{}}
        \textcolor{textmuted}{Tutor scaffolding} & \textcolor{accentenv}{\textbf{True}} \\
        \textcolor{textmuted}{Tutor on topic} & \textcolor{accentenv}{\textbf{True}} \\
        \textcolor{textmuted}{Student answers correctly} & \textcolor{accentenv}{\textbf{True}}
    \end{tabular}
    }
};


\draw[flowline] (state.east) -- (policy.west) 
    node[midway, above, font=\scriptsize\sffamily, text=textmuted] {};

\draw[flowline] (promptpool.south) -- (policy.north) 
    node[midway, right, font=\scriptsize\sffamily, text=textmuted] {};

\draw[flowline] (policy.east) -- (action.west) 
    node[midway, above, font=\scriptsize\sffamily, text=textmuted] {};

\draw[flowline] (action.south) -- (environment) 
    node[midway, right, font=\scriptsize\sffamily, text=textmuted] {};

\draw[flowline] (environment.west) -- (evaluator.east) 
    node[midway, above, font=\scriptsize\sffamily, text=textmuted] {Transcript};

\draw[rewardline] (evaluator.north) -- ++(0, 1.5) coordinate (turn1) -| (policy.south);
\path (evaluator.north) -- (turn1) node[midway, left=2pt, font=\scriptsize\sffamily\bfseries, text=mochamousse] {Score ($y \in \mathcal{Y}$)};

\end{tikzpicture}
    \caption{The tutoring system encodes subject-specific input ($X$) to select a tutoring prompt ($p$) from a pool of pedagogical strategies ($P$). After simulating a multi-turn conversation, an LLM-based evaluator decomposes the interaction into 14 pedagogical criteria to compute a feedback signal ($Y$).}
    \label{fig:framework}
\end{figure}
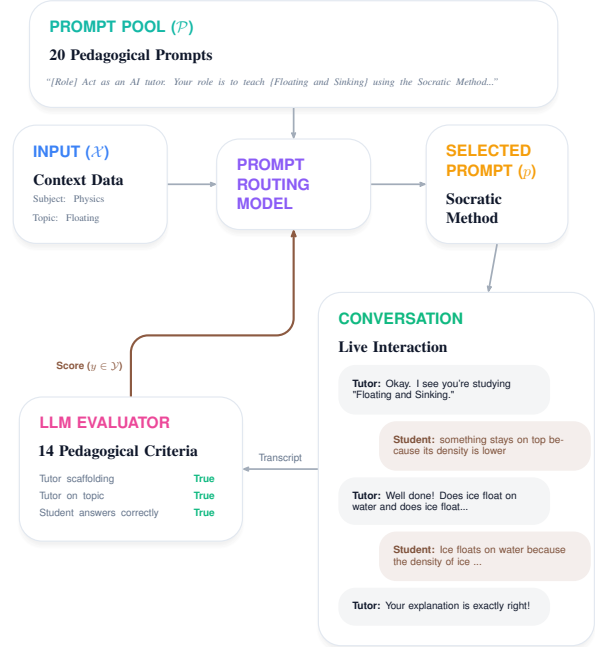

Furthermore, current approaches rely on static prompting \citep{gupta2025beyond} for a single subject (mathematics) but do not adapt to the diverse academic disciplines (such as French, math, or geography). While automatic prompt engineering \citep{Zhou2022LargeLM} attempts to optimize instructions, it typically converges on a globally static prompt. Additionally, pre-trained semantic spaces often suffer from representation degeneration \citep{gao2019representation}, hindering the current routing model from distinguishing between different subjects. To overcome these limitations, we propose an adaptive, subject-aware prompt routing framework capable of adjusting instructional strategies across subjects.

However, to successfully train and optimize such an adaptive routing system, a core challenge for LLM-based tutoring systems lies in designing a robust measure of pedagogical quality that enables an adaptive LLM-tutor to bridge the gap between the ideal world of simulations and sparse real-world interactions. Additionally, in real-world scenarios, users often skip formal exercises after interacting with the LLM tutor, resulting in a lack of immediate feedback for model training. Previous works rely on a single black-box score \citep{10.5555/3666122.3668142}, but our LLM evaluator decomposes instructional efficacy into 14 observable criteria grounded in educational theory (e.g., scaffolding) \citep{wood1976role}.
By calibrating these feedback features against human labels, our score function serves as a grounded proxy for learner success, even when the explicit exercise score is missing.

%
%

We propose a Subject-aware adaptive tutoring system, based on LLM evaluator feedback (Figure~\ref{fig:framework}). We train a prompt routing model with a simulator of student dialogue. Then we deploy our system in a real-world test with high school students. 
%
%
To evaluate our framework, we employ independent t-tests for simulation metrics and the Mann-Whitney U test \citep{doi:https://doi.org/10.1002/9780470479216.corpsy0524} for real-world downstream impact. Our key contributions are:
\begin{enumerate}
    \item We introduce an \textit{LLM evaluator} that serves as an immediate proxy for learners' success, even in the absence of downstream exercise scores. 
    Moreover, this proxy is positively associated with the exercise conversion of real students ($0.599$ with converted vs. $0.560$ with not converted, $N=656$, $p=0.037$).
    \item We formulate adaptive tutoring as a \textit{prompt routing} problem, using topic and subject information as input. 
    The router can specialize pedagogical strategies across subjects, outperforming two static baselines in simulation ($0.694$ vs. $0.647$ and $0.64$, $p<0.001$).
    \item We validate our approach 
    in a \textit{real-world} setting with Dutch high-school student interactions ($N=656$). The greedy router is comparable with our chosen baseline in the conversion rate ($19.1\%$ vs. $19.6\%$), but a stochastic router achieves a higher conversion ($28.1\%$). Although this downstream improvement does not reach statistical significance, it underscores both the potential of dynamic routing and the ongoing challenges of sim-to-real transfer.
\end{enumerate}


\section{Related Work}
\paragraph{Adaptive Tutoring and Prompt Routing}
Large language models show significant potential in personalized education \citep{malik2025scaffolding, electronics14234683}. While early prompt optimization focused on a global optimum \citep{Zhou2022LargeLM, yang2023large}, and recent education applications often rely on static prompts for a single subject like mathematics \citep{gupta2025beyond}. Moreover, recent approaches model this as sequential decision-making \citep{nam2025efficient, dinucu-jianu-etal-2025-problem} by using reinforcement learning (RL).

Meanwhile, routing has emerged as a strategy in LLM inference. While most existing work focuses on model routing, selecting between LLMs to optimize the cost-performance trade-off \citep{chen2023frugalgptuselargelanguage, ong2025routellmlearningroutellms} or utilizing bandit-based method for efficient inference \citep{li2025llmbanditcostefficientllm, jitkrittum2025universalmodelroutingefficient}, recent studies demonstrate that prompt routing can enhance task-specific accuracy, such as in arithmetic reasoning \citep{repec:bcp:journl:v:10:y:2026:i:19:p:605-611}.

We combined these paradigms to formulate an adaptive pedagogical strategy as a prompt routing problem, solved using a contextual bandit framework.
Our subject-aware router dynamically navigates pedagogical strategies (e.g., scaffolding vs.\ deep reasoning).
\paragraph{AI Feedback and Feature-based Reward Learning}
While AI feedback offers a scalable alternative to costly human feedback \citep{bai2022constitutional}, conventional scalar scores are readily affected by high variance and reward gaming \citep{amodei2016concrete, skalse2022defining}. 
Inspired by feature-based reward learning in human cognition \citep{schultner2025feature}, we constructed 14 binary educational features (e.g., understanding, curiosity), instead of a single black-box score.

\paragraph{AI Tutor Evaluation Frameworks}
Recent works have proposed comprehensive evaluation taxonomies for AI tutors, such as \citet{maurya-etal-2025-unifying} and \citet{2024arXiv240712687J}. While these established frameworks provide rigorous offline metrics to benchmark static or fine-tuned LLMs, they are computationally heavy and not designed for real-time training loops. Instead, our 14-feature evaluator is specifically designed to be lightweight, enabling immediate feedback to drive dynamic, on-policy prompt routing during live interactions.

\paragraph{Bridging the Sim-to-Real Gap}
An important challenge in LLM-based tutoring systems is the sim-to-real gap \citep{zhou2026mindsim2realgapuser}, since the routing model may overfit to idealistic feedback in simulation \citep{gao2023scaling, singhal2023learning}. This misalignment is critical in education, as simulators often exhibit sycophancy, unlike real students \citep{perez2022discover}. 
Recent work emphasizes the reward calibration \citep{mao-etal-2024-dont} to ensure stable alignment in deployment. 
Following robust decision-making \citep{moos2022robust}, we implement a score smoothing mechanism to align the score distributions between simulated environments and real-world deployment. 

\section{Subject-aware Adaptive Tutoring Framework}\label{sec:methods}
We formalize adaptive prompt selection as a contextual bandit routing problem, as a tuple $\langle \mathcal{X}, \mathcal{P}, \mathcal{Y} \rangle$. The router observes a pedagogical input $x \in \mathcal{X}$, and selects a tutoring prompt from the prompt pool $p \in \mathcal{P}$. The environment then returns an evaluative score $y \in \mathcal{Y}$ to optimize the router (Figure~\ref{fig:framework}).

\subsection{Input Representation and Prompt Space}
\label{subsec:state-action}
\paragraph{Subject-aware Input ($\mathcal{X}$)} 
To enable cross-subject decision-making while mitigating representation degeneration (anisotropy) observed in pre-trained embeddings \citep{gao2019representation}, we utilize a hybrid input representation. The input $x$ fuses a 768-dimensional topic embedding from an E5 encoder ($\mathbf{e}_{topic}$) \citep{Wang2022TextEB} with a learnable 64-dimensional subject ID embedding ($\mathbf{e}_{subject}$). In Equation~\ref{eq:x}, $\text{LN}$ denotes Layer Normalization, and $\oplus$ is concatenation. This dual-path architecture explicitly injects subject boundaries, effectively unfolding the collapsed latent space (App.~\ref{app:anisotropy}).
\begin{equation}
\label{eq:x}
\mathbf{x} = \text{LN}(f_{\phi}(\mathbf{e}_{topic}) \oplus f_{\psi}(\mathbf{e}_{subject}))
\end{equation}

\paragraph{Prompt Space ($\mathcal{P}$)} 
The prompt space comprises 20 pedagogical prompts with different instructional strategies (e.g., Socratic scaffolding, Feynman role reversal). This space size was chosen to provide a diverse exploration space for the routing model to adapt to different educational subjects. These prompts are verified by educational experts and directly originate from the foundational baseline prompt currently deployed in the production environment. The foundational prompts and their strict constraints (e.g., the `2-attempt rule') were explicitly authored by human educational experts. We utilized LLMs to increase variation (App.~\ref{app:prompt_pool}) based on expert-crafted templates (App.~\ref{app:static_baseline}). 

\subsection{Evaluative Score Calibration ($\mathcal{Y}$)}
\label{subsec:reward}
As evaluator LLM, we used GPT-5-mini \citep{singh2025openai} with \emph{medium} reasoning effort and a temperature setting of 1. The pedagogical score $y$ is a weighted sum of 14 binary pedagogical features (App.~\ref{app:rubrics}), denoted as $c_i \in \{0, 1\}$, $y_{raw} = \sum_{i=1}^{14} w_i \cdot c_i$. The weights $w_i$ are derived from Pearson correlations with human-labeled learning outcomes (App.~\ref{app:correlation}). To ensure robustness of feedback features, we aggregate 3 independent evaluation runs via majority voting, improving self-consistency to $94.31\%$ (App.~\ref{app:consistency}). 

Furthermore, to prevent the routing model from over-optimizing on the behaviors of simulated students, we fix the distributional shift between simulation and reality by applying a sigmoid calibration: $y_{calibrated} = [1 + e^{-K \cdot (y_{raw} - 0.5)}]^{-1}$. Through Jensen-Shannon (JS) divergence analysis against empirical distributions, we fix $K=3.0$ to bridge the sim-to-real gap (App.~\ref{app:calibration}).

\subsection{Residual Routing Architecture and Optimization}
\label{subsec:policy_opt}
For the routing model, the Actor-Critic network ($\pi_\theta, V_\omega$) employs a residual subject injection mechanism, (App.~\ref{app:policy_arch}). To prevent the categorical subject anchors ($\mathbf{e}_{subject}$) from being diluted through deep non-linear transformations, we implement a skip-connection injecting $\mathbf{e}_{subject}$ directly into the final output layers. Formally, the Actor forward pass is:
\begin{equation}
\pi(\cdot|\mathbf{x}) = \text{Softmax}(W_{out} [h_2 \oplus \mathbf{e}_{subject}] + b_{out})
\end{equation}
where $h_2$ is the output of a 2-layer MLP processing the hybrid input $\mathbf{x}$. This ensures that subject-specific requirements strongly govern the output distribution. We initialize $W_{out}$ with a low standard deviation ($0.01$) for initial exploration.

We optimize the prompt selection $\pi_\theta$ using PPO \citep{schulman2017proximal}. While PPO is traditionally applied for multi-step RL, restricting the episode length to a single step adapts it into an effective contextual bandit algorithm, where the actor-critic architecture naturally provides baseline-subtracted policy gradient updates. We augment it with Experience Replay \citep{mnih2015human} and importance sampling corrections to allow multiple optimization epochs on historical transitions $(x, p, y, \pi_{old})$, ensuring sample efficiency despite the high latency of LLM environments.

\section{Experimental Setup}
\label{sec:experiments}
To safely optimize the routing model and observe its behavior without risking real-world pedagogical failures, we developed a controlled simulation environment. In this section, we detail this simulator alongside our evaluation baselines, metrics, and implementation settings.

\subsection{Simulation Environment}
\label{subsec:simulation}
In our simulation, we modeled three distinct simulated-student profiles: Motivated, Mediocre, and Unmotivated. These profiles were derived from real transcripts and validated by educational experts. To ensure behavioral realism, 
while preventing LLM verbosity, we extracted stylistic summaries and injected stochastic ``inner monologues'' (e.g., \texttt{[Think: you are confused]}) to guide the student’s reasoning (App.~\ref{app:student_persona}). Representative samples of simulated interactions 
are provided in App.~\ref{app:sim_conversation} for qualitative inspection. During training, sessions sample a curriculum topic and run for 1 to 15 turns, depending on the student's motivation. The environment executes the selected strategy $p \in \mathcal{P}$, generating session-level transitions $(x, p, y)$ to provide stable routing signals.

\subsection{Baselines and Evaluation}
\label{subsec:eval_method}
We evaluated our dynamic routing model \begin{enumerate*}[label=(\arabic*)]
    \item in simulation against various baselines, and
    \item in a real-world setting compared to a fixed baseline. 
\end{enumerate*}

\paragraph{Simulation Benchmark} To evaluate pedagogical efficacy in a controlled setting, we compared our dynamic routing against two baselines: (1) \textit{Production Baseline}: the expert-refined prompt currently deployed by a tutoring platform from Futurewhiz. (2) \textit{Literature Baseline}: a static adaptation of the pedagogical prompt proposed by \citet{dinucu-jianu-etal-2025-problem}, integrated with dynamic topic placeholders (App.~\ref{app:dinucu_prompt}).
To ensure a fair comparison within our discrete prompt routing framework, we provide a single system instruction, disabling dynamic strategy switching during interactions.

\paragraph{Real-World A/B Testing} We conducted an online A/B test, randomly assigning existing students at the user-level across multiple sessions to either the static production baseline or the adaptive router.  
The live deployment naturally covered 13 high-school subjects driven by real user traffic.
The router consists of two operation modes: exploitation and exploration.
We differentiate between these two modes because the exploitation is expected as an optimal solution, stabilizing real-user experience, while the exploration can ensure diverse training data.
Moreover, we filtered for sessions with $turns > 1$, resulting in $N=656$ valid interactions with 359 students. Our experiments ran in March and April 2026.


Beyond the pedagogical score, we pair interaction logs with same-day exercise results to measure the educational impact. Our metrics include the conversion rate (the proportion of sessions transitioning to formal practice) and exercise accuracy, and we use the Mann-Whitney U test \citep{doi:https://doi.org/10.1002/9780470479216.corpsy0524} to determine statistical significance. To isolate the pedagogical effect from individual learner level, we conducted a within-user analysis for accuracy, comparing sessions where the AI achieved above-average scores against below-average ones for the same student.  

\subsection{Implementation Details}
\label{subsec:implementation}
\paragraph{Training}
For simulation benchmarking, the routing model was trained in simulation for 16,000 steps, using the AdamW optimizer \citep{loshchilov2017decoupled} and PPO with Experience Replay. This extended training allows us to observe the router's evolution 
in a controlled environment.

For live deployment, we restricted the initial simulation pre-training to 2,000 steps and finetuned the router using the data from the Real-World A/B test. This strategy provides a warm start and prevents the router from overfitting on the synthetic data. The training data is collected daily from real-user interaction and updates the routing model. Moreover, to prevent the routing model from developing a bias toward high-traffic subjects, we use a balanced sample for the simulation and live deployment data. 

\paragraph{Inference}
The router estimates a probability distribution over teaching strategies. 
In live deployment, we use the router for 80\% of the time in \emph{exploitation} mode, greedily taking the strategy with the highest probability, as we assume it to be the optimal, ensuring user experience. The other 20\% of the time, we perform \emph{exploration} by weighted sampling according to the estimated probabilities, to diversify our training data and provide the necessary on-policy actions for continuous training.

\paragraph{Hyperparameter Setting}
App.~\ref{app:hyperparameters} shows our hyperparameter settings for simulation training. However, for the live deployment, we reduced the batch size from 64 to 16 to accommodate the sparsity of real data. Additionally, to tackle the inconsistency in the amount of daily incoming data, we dynamically adjust the experience replay ratio between 1:1 and 1:3, updating the model when sufficient data is available, ensuring frequent model updates.



\section{Experiment Results}\label{sec:results}
In our experiments, we sought to answer research questions on (1) score validity and discriminative power, (2) adaptive prompt learning in simulation, and (3) sim-to-real impact.

\subsection{Score Validity and Discriminative Power}
\label{subsec:rq1}
\begin{figure}[tb]
    \centering
    \includegraphics[width=.9\linewidth]{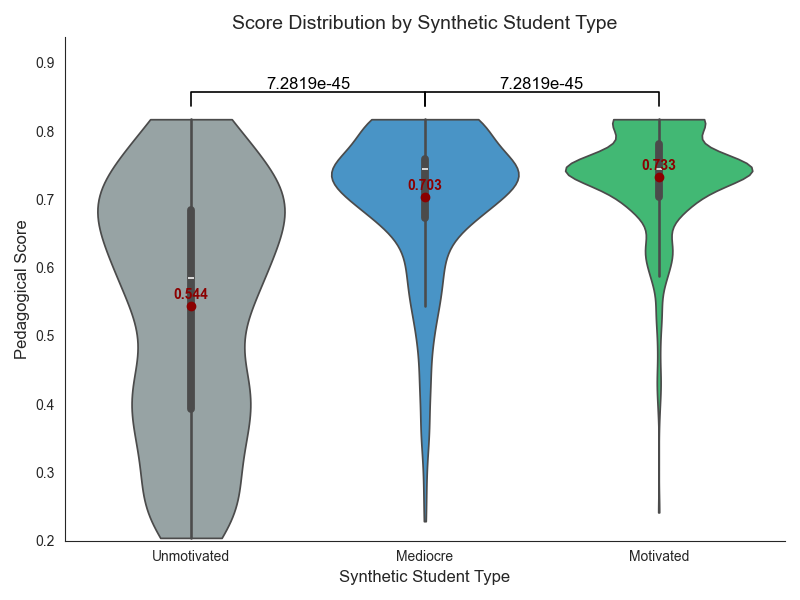} 
    \caption{Score Distribution by Simulated-Student Profile. The clear separation between ``Motivated'', ``Mediocre'', and ``Unmotivated'' simulated-student profiles establishes a difficulty gradient for the 
    model and confirms the LLM evaluator's discriminative validity.}
    \label{fig:student_type}
\end{figure}

To validate AI feedback for training the pedagogical features, we evaluate the interaction's pedagogical quality across student profiles in simulation. As illustrated in Figure~\ref{fig:student_type}, the score distribution across all benchmarking methods exhibits a clear difference. Motivated students yield the highest mean score (0.733), while unmotivated students result in a lower mean (0.544) with a wider distribution.

Statistical significance was confirmed via independent t-tests, showing highly significant differences between all groups ($p < 0.001$), confirming that the LLM evaluator successfully discriminates between simulated-student states. 

\begin{table*}[tb]
    \centering
    \caption{Comparison of Empirically Optimal Strategies vs. Model-Predicted Strategies across Subjects in simulation. The ``Empirical Best Prompt'' is the prompt that yielded the highest average pedagogical score in the entire data. The ``Model Predicted Prompt'' is the router's output at the end of training. This comparison shows that the router can explore and select better strategies based on different subjects. The metric ``Gap'' is the regret between the theoretical maximum score and the model's chosen prompt.}
    \label{tab:empirical_vs_model_sim}
    \resizebox{\textwidth}{!}{
    \begin{tabular}{llclcc}
        \toprule
        \textbf{Subject} & \textbf{Empirical Best Prompt} & \textbf{Mean Score} & \textbf{Model Predicted Prompt} & \textbf{Predicted Mean Score} & \textbf{Gap} \\
        \midrule
        Geography & \textbf{Analytical\_opt} & 0.712 & \textbf{Analytical\_opt} & 0.712 & \textbf{0.000} \\
        Biology       & Analytical      & 0.696 & Analytical\_opt & 0.674 & 0.022 \\
        Economics       & \textbf{Analytical\_opt} & 0.698 & \textbf{Analytical\_opt} & 0.698 & \textbf{0.000} \\
        English        & Challenge       & 0.721 & Analytical\_opt & 0.716 & 0.005 \\
        French          & \textbf{Feynman}         & 0.722 & \textbf{Feynman}         & 0.722 & \textbf{0.000} \\
        History   & Best            & 0.723 & Analytical\_opt & 0.710 & 0.013 \\
        PhysChem          & \textbf{Analytical\_opt} & 0.703 & \textbf{Analytical\_opt} & 0.703 & \textbf{0.000} \\
        Physics    & \textbf{Analytical}      & 0.727 & \textbf{Analytical}      & 0.727 & \textbf{0.000} \\
        Dutch     & Step            & 0.693 & Analytical\_opt & 0.654 & 0.039 \\
        Calculus        & \textbf{Analytical\_opt} & 0.685 & \textbf{Analytical\_opt} & 0.685 & \textbf{0.000} \\
        Learning Skills     & Emotion\_First  & 0.722 & Analytical\_opt & 0.709 & 0.012 \\
        Language           & \textbf{Analytical\_opt} & 0.713 & \textbf{Analytical\_opt} & 0.713 & \textbf{0.000} \\
        Mathematics       & \textbf{Feynman}         & 0.705 & \textbf{Feynman}         & 0.705 & \textbf{0.000} \\
        \midrule
        \multicolumn{6}{l}{\textbf{Strategy Alignment Rate (Gap $\le$ 0.01): 69.2\%}} \\
        \bottomrule
    \end{tabular}
    }
\end{table*}
\subsection{Adaptive Prompt Learning in Simulation}
\label{sec:ppo_training}
\begin{figure*}[tb]
    \centering
    \begin{subfigure}{0.38\textwidth}
        \centering
        \includegraphics[width=\linewidth]{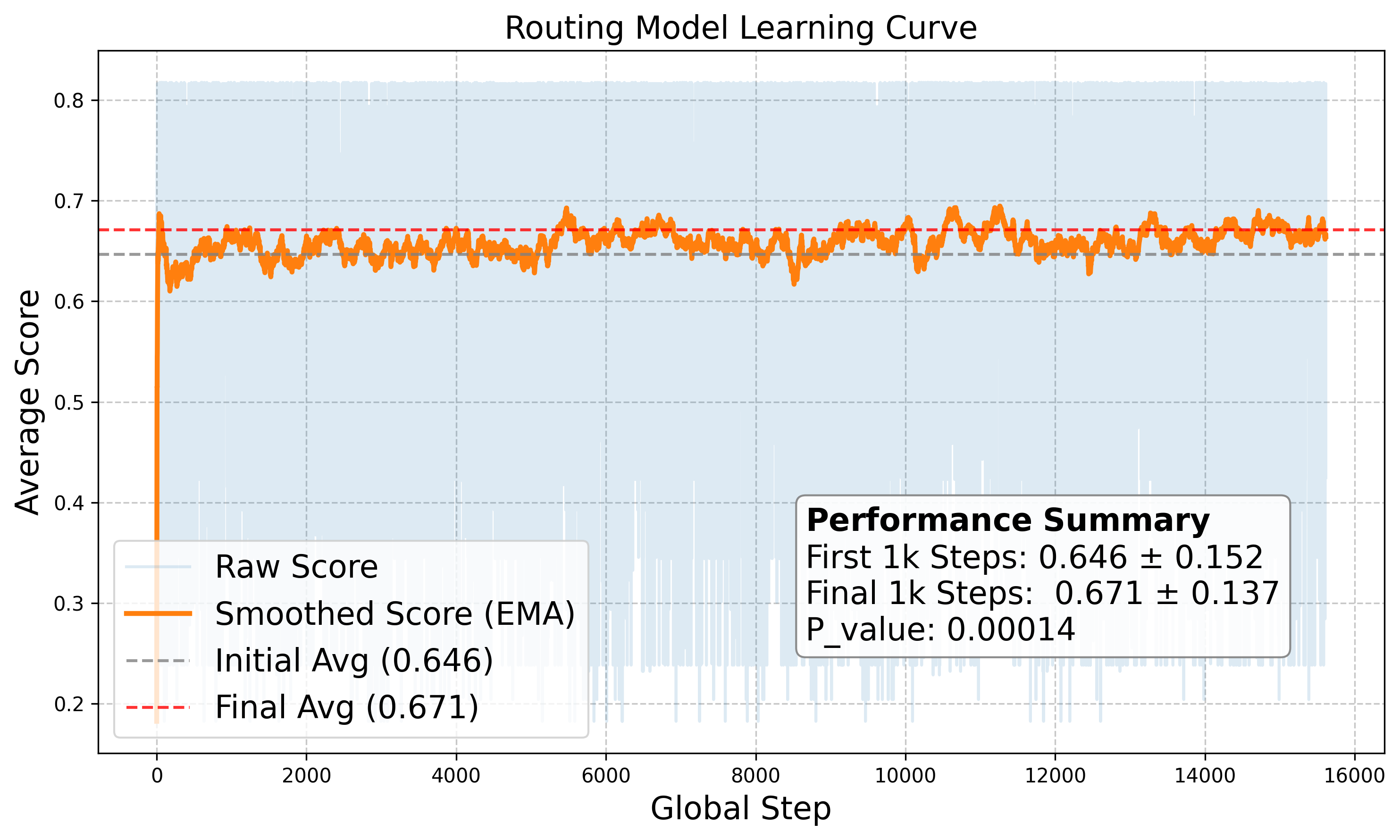}
        \caption{Overall Convergence}
        \label{fig:lc_overall}
    \end{subfigure}
    \hfill
    \begin{subfigure}{0.58\textwidth}
        \centering
        \includegraphics[width=1\linewidth]{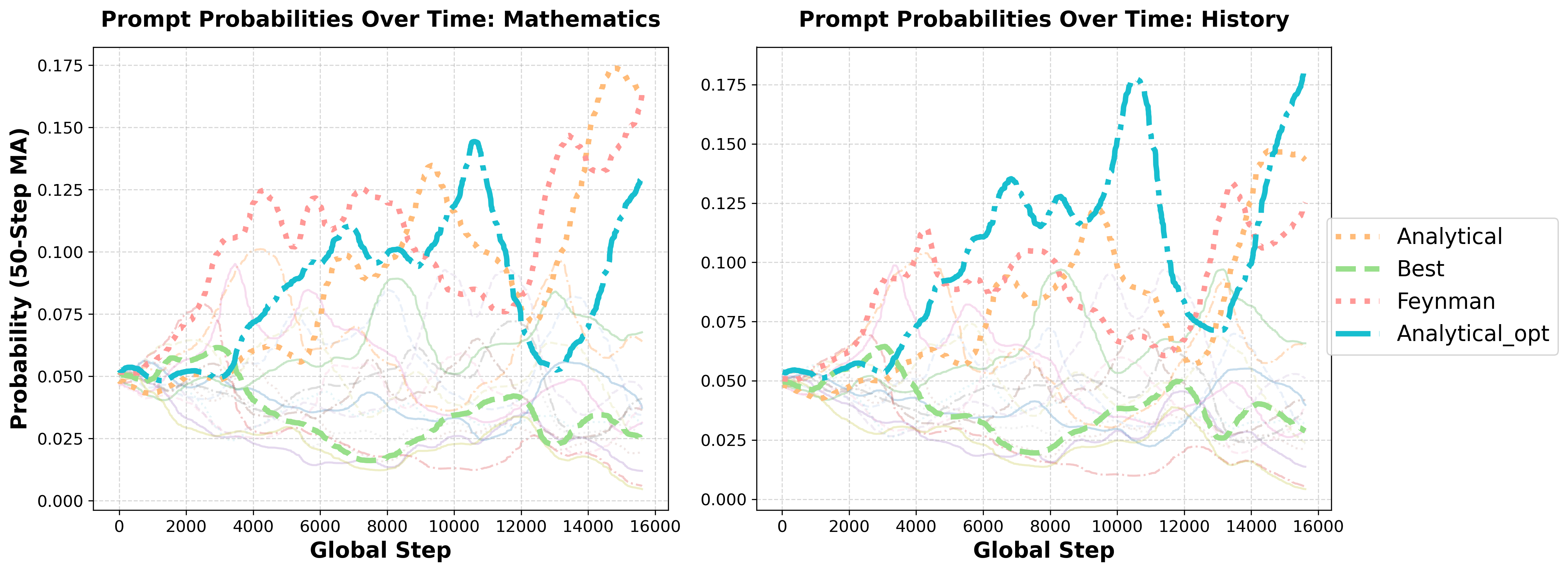}
        \caption{Subject-specific Adaptation  (Mathematics vs. History)}
        \label{fig:prob_comparison}
    \end{subfigure}
    
    \caption{Learning dynamics in simulation. (a) Stable score growth despite high environment noise. (b) Strategy evolution showing clear crossovers in Math versus local optima traps in History. Shared legend: \textit{Best} (green), \textit{Feynman} (pink), \textit{Analytical} (orange), \textit{Analytical\_opt} (teal).}
    \label{fig:combined_results}
\end{figure*}

For our second research question (Figure~\ref{fig:lc_overall}) the router successfully converges from an initial mean score of 0.646 to 0.671 ($+2.5\%$) with reduced variance after 16,000 steps. Despite noise from unpredictable student behavior, the routing model elevates performance and achieves statistically significant gains in score stability. Furthermore, in the simulation benchmark (App.~\ref{app:benchmark_results}), the routing model outperforms both the \textit{Production} and \textit{Literature} baselines ($0.694$ vs. $0.647$ and $0.64$, $p<0.001$), largely by eliminating the tail of poor interactions (scores $< 0.4$).

To quantitatively validate whether the router discovered the optimal
Subject-Prompt mapping, we conducted an empirical alignment analysis. We compared the
model’s final predicted prompt against the empirical ground truth (Table~\ref{tab:empirical_vs_model_sim}), which was determined by calculating the maximum average 
score across all available interaction logs for that subject. 
The results demonstrate a high degree of strategy alignment. In 8 out of the 13 evaluated subjects, the routing model achieved a perfect match (Gap = 0.000), successfully adopting the statistically optimal pedagogical strategy. In subjects where a gap exists, the regret margins remain small (averaging under 0.02 among non-matching subjects). This shows that there is no ``silver bullet'' strategy for all disciplines.

Subject-aware adaptation is evidenced by investigating the routing predictions during training across disciplines (Figure~\ref{fig:prob_comparison}). By utilizing subject anchors to mitigate representation degeneration (Section~\ref{subsec:state-action}), the dual-path architecture prevents mode collapse. In Mathematics, the model successfully transitions from global priors to the subject-optimal \textit{Feynman} strategy.

However, in subjects like History (see Table~\ref{tab:empirical_vs_model_sim} and Figure~\ref{fig:prob_comparison}), the routing model converges to \textit{Analytical\_opt}, instead of the empirically superior \textit{Best} strategy. This reflects a risk-avoiding behavior in our routing model; when potential gains are marginal ($<0.013$), the router prioritizes score stability over exploration. This confirms that the framework effectively maps disciplinary contexts to near-optimal interventions, providing higher instructional value across the curriculum.
\begin{figure}[tb]
    \centering
    \includegraphics[width=.9\linewidth]{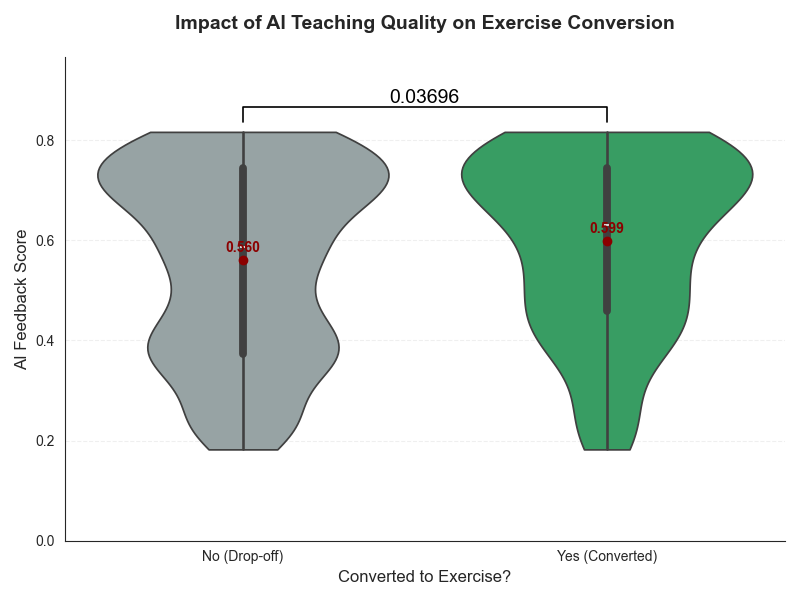}
    \caption{Comparison of pedagogical score between converted and non-converted sessions. Successful conversions to practice correlate with higher feedback scores ($0.599$ vs. $0.560$, $N=656$, $p=0.037$).}
    \label{fig:comparison_conversion_reward}
\end{figure}
%

\begin{figure*}[tb]
    \centering
    \begin{subfigure}{0.49\textwidth}
        \centering
        \includegraphics[width=\linewidth]{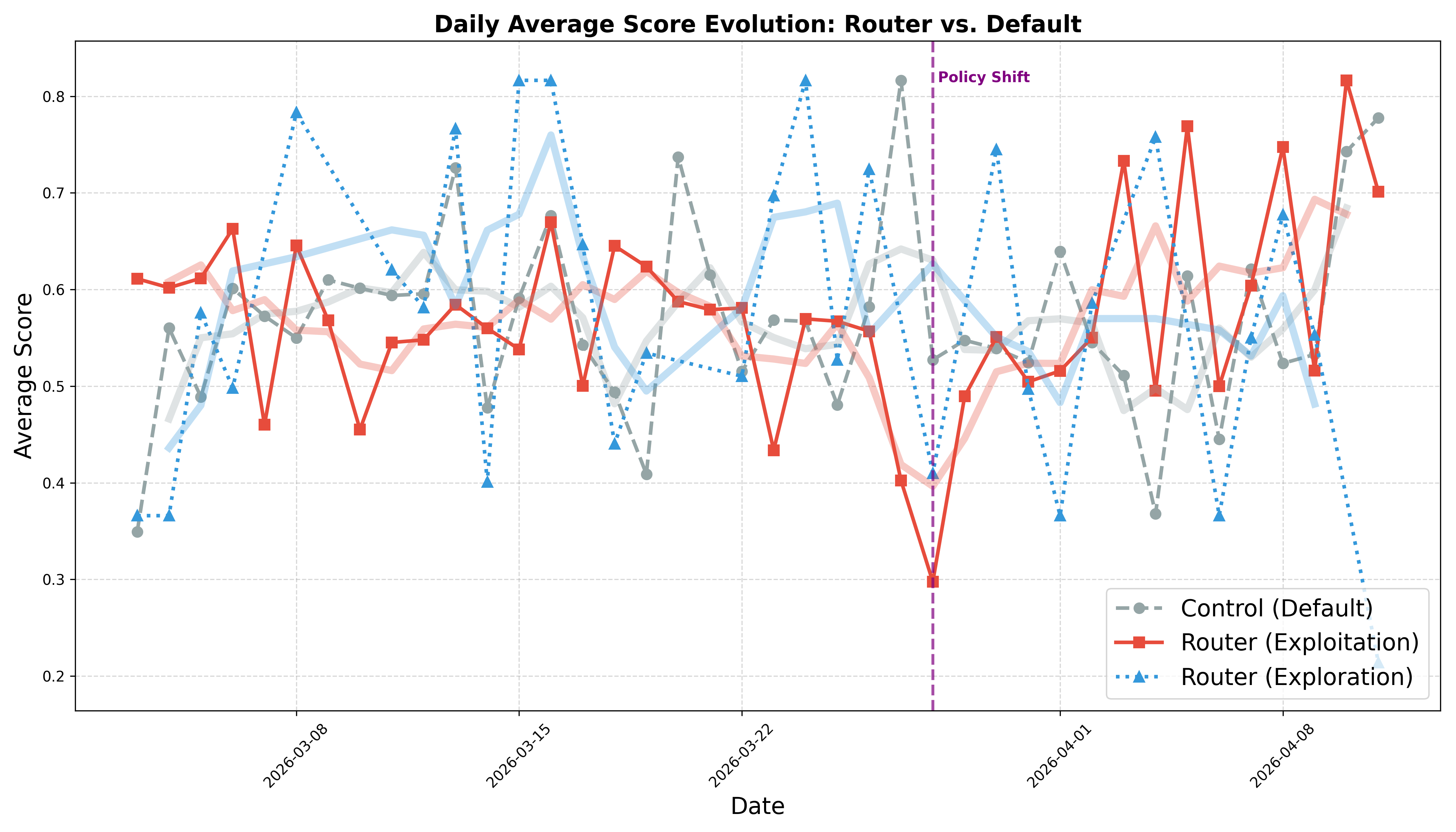}
        \caption{Daily Average Score Evolution}
        \label{fig:daily_reward}
    \end{subfigure}
    \hfill
    \begin{subfigure}{0.49\textwidth}
        \centering
        \includegraphics[width=\linewidth]{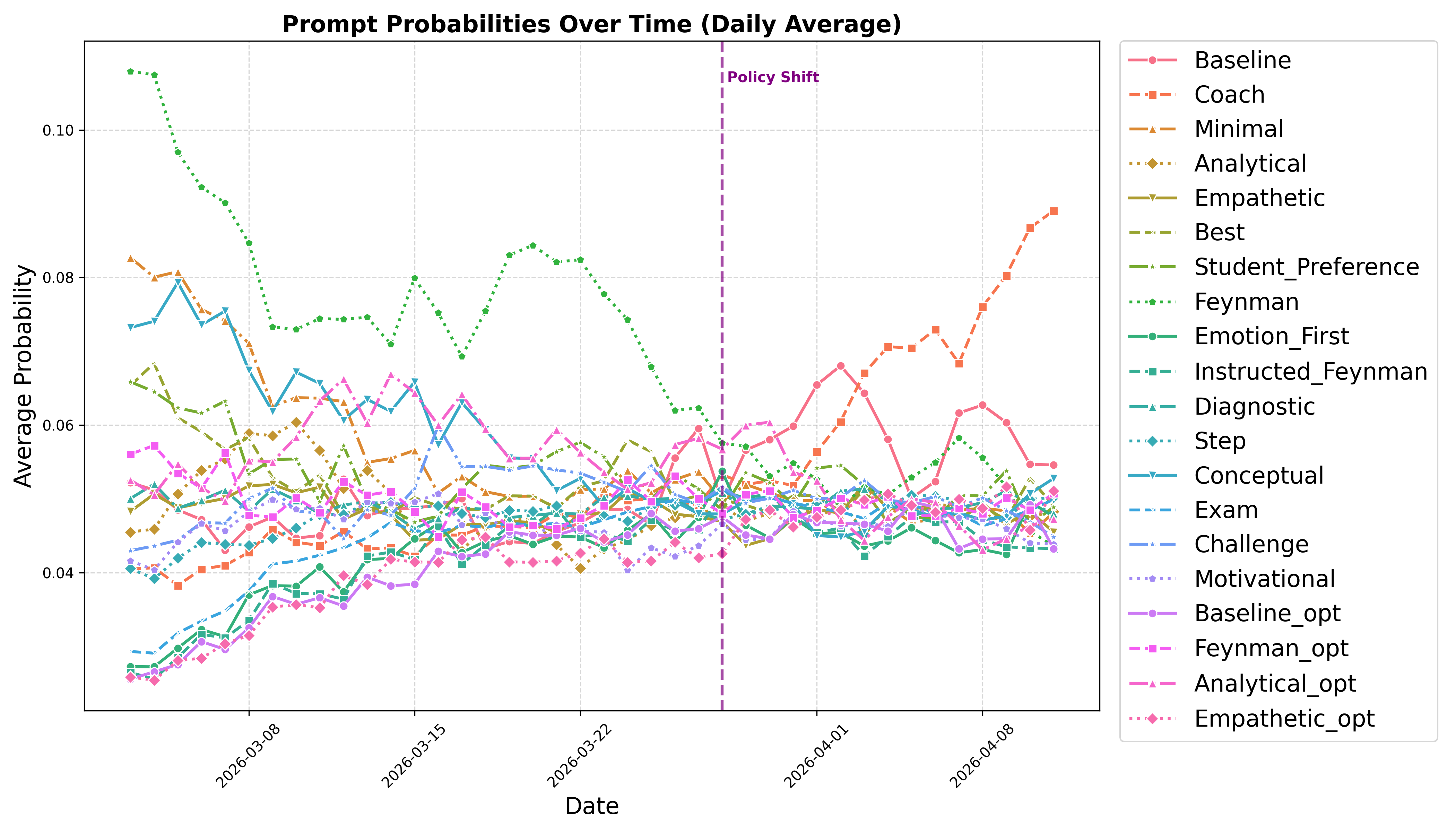}
        \caption{Prompt Probabilities (Daily)}
        \label{fig:daily_prob}
    \end{subfigure}
    
    \caption{Training score trajectory during the live deployment phase. The ascent after March 28 suggests the transition from Exploration to Exploitation, where the router successfully converges on the \textit{Coach} strategy. This demonstrates the router’s ability to refine its pedagogical strategy through real-world feedback, surpassing the production baseline and the preferences observed in simulation (\textit{Feynman}).}
    \label{fig:policy_evolution_combined}
\end{figure*}

\subsection{Real-World Deployment and Educational Impact}
We performed a live A/B test with $N=656$ interactions with 359 Dutch high-school students. When evaluated over the entire deployment period, the router maintained robust instructional standards ($0.565$ vs.\ baseline $0.569$, $p=0.635$). 
In the final 7 days of deployment, the router achieved a higher average score of $0.612$ ($n=35$) compared to the baseline's $0.596$ ($n=51$), while $p=0.461$.

As shown in Table~\ref{tab:online_strategy_comparison_full} (App.~\ref{app:emp_real}), the converged model achieved an 80.0\% alignment rate for high-confidence subjects ($n \ge 5$), with small gap ($0.008$) in complex domains like Mathematics. However, data sparsity occasionally induced safe behavior. E.g., in History ($n=4$), the sparse real data caused the routing model to retreat to the globally stable \textit{Coach} rather than committing to the optimal \textit{Baseline} strategy. Beyond alignment, the routing model enhanced instructional efficiency, guiding students through many interactions ($\text{turns} \ge 5$) in fewer turns ($N=381$, $p=0.0073$; App.~\ref{sec:ab_test}). While this reduction in conversational turns remains visible across the unfiltered dataset, the difference does not reach statistical significance ($p=0.63$) due to the high variance from short, noisy interactions.

Those improvements stem from the router decoupling simulation biases (Figure~\ref{fig:policy_evolution_combined}). Initially favoring the deeper reasoning \textit{Feynman} strategy from simulation, the strategy switched to the scaffolding \textit{Coach} strategy after live feedback. This shift aligned the strategy with student needs, yielding superior live performance ($p=0.0253$, App.~\ref{app:emp_real}).

Finally, we linked AI pedagogical quality to downstream student behavior. The analysis (Figure~\ref{fig:comparison_conversion_reward}) shows that the higher pedagogical scores are associated with successful exercise conversion ($0.599$ with converted vs. $0.560$ with not converted, $N=656$, $p=0.037$). 
Nevertheless, the system-level comparison of the conversion rate (App.~\ref{app:comp_conversion_real}) shows the exploitation router is comparable with the static baseline ($19.1\%$ vs. $19.6\%$). Interestingly, the exploration router yields a higher conversion rate $28.1\%$. Rather than exploiting an (in simulation) optimal routing policy, sampling stochastically is linked to a higher conversion rate. 
Furthermore, the subsequent within-user exercise accuracy differences were not statistically significant ($0.76$ vs. $0.70$, $N=56$, $p=0.281$, App.~\ref{app:comp_accuracy}).

Beyond the quantitative improvements, qualitative analysis of real-world interactions (App.~\ref{app:real_chat}) confirms the router's pedagogical efficacy. The router-selected \textit{Coach} prompt has robust scaffolding, handling real-world noise (e.g., student typos) and autonomously executing level-adaptation.

\section{Conclusion}
We present a Subject-aware approach for adaptive pedagogical prompting. We use a human-aligned score mechanism to train a routing model that adapts instructional strategies to specific subjects. In an experiment with 359 high-school students, 
we can use higher scores as an index for exercise conversion.

We first validate that decomposing instructional efficacy into 14 observable criteria yields a robust score that aligns with human pedagogical standards. 
Second, our results demonstrate that the router raises the lower bound of the feedback score, compared to static baselines, and achieves high subject alignment with empirical optima in simulation (69.2\%).
Finally, in real-world deployment, the routing model improves instructional efficiency, reducing deep interactions by around 3 turns ($p=0.007$) in the longer conversations, and leads to an increase in conversion rate when deployed. Although the observed increase in the exercise conversion rate does not reach statistical significance, the system demonstrates a clear potential to optimize instructional strategies toward better learner engagement.

Future work can refine the topic embeddings for topic-level specialization, balance high-traffic subjects that dominate learning, 
or focus on updating the router to include conversation history, or by performing $\epsilon$-greedy search to improve strategy selection outcomes. Moreover, exploring an approach with continuous prompt space could further yield hybrid pedagogical strategies, providing greater flexibility in decision-making. Furthermore, by introducing Epistemic State Specification (ESS) \citep{yuan2026validstudentsimulationlarge}, the simulator would capture more complex cognitive nuances of real students.

\section*{Limitations}
We identify several limitations in our study. 

First, many students did not proceed to formal exercises right after tutoring, making it difficult to systematically evaluate exact learning outcomes. This limited our ability to measure immediate knowledge gain through test results. Future longitudinal studies with a high student test conversion rate are needed to address this.

Second, the training and deployment processes highlighted the challenge of domain imbalance. Although we applied balanced sampling during simulation, we observed that the router develops a strong bias toward pedagogical strategies that perform well across the most common subjects. This domain-independent strategy, coupled with the inherent real-world data sparsity in long-tail subjects (e.g., Economics or Physics), makes it difficult for the policy to confidently differentiate and converge on niche, subject-specific optimal strategies within a limited live-deployment window.

Third, our framework relies on a discrete prompt pool. While this ensures pedagogical safety and prevents harmful AI behaviors, it prevents the model from generating novel or hybrid instruction strategies.

Finally, the data sparsity is a critical bottleneck in our real-world deployment, referring the most significant challenge to an effective sim-to-real transfer. This limitation not only constrains the statistical power of our downstream analyses but also slows the online convergence of the routing model. 

\section*{Ethics Statement}
All student conversation transcripts were processed within a secure, closed-loop environment by a professional high-school tutoring company. Data were handled following the General Data Protection Regulation (GDPR), and access was restricted to the research team. Explicit consent for data usage for research purposes was obtained through agreements with the participating educational platform.

\bibliography{anthology,custom}
\bibliographystyle{acl_natbib}

\appendix
\onecolumn

\section{Architecture of the Actor-Critic Routing Model}
\label{app:policy_arch}
\begin{figure}[H]
    \centering
    \includegraphics[width=.7\linewidth]{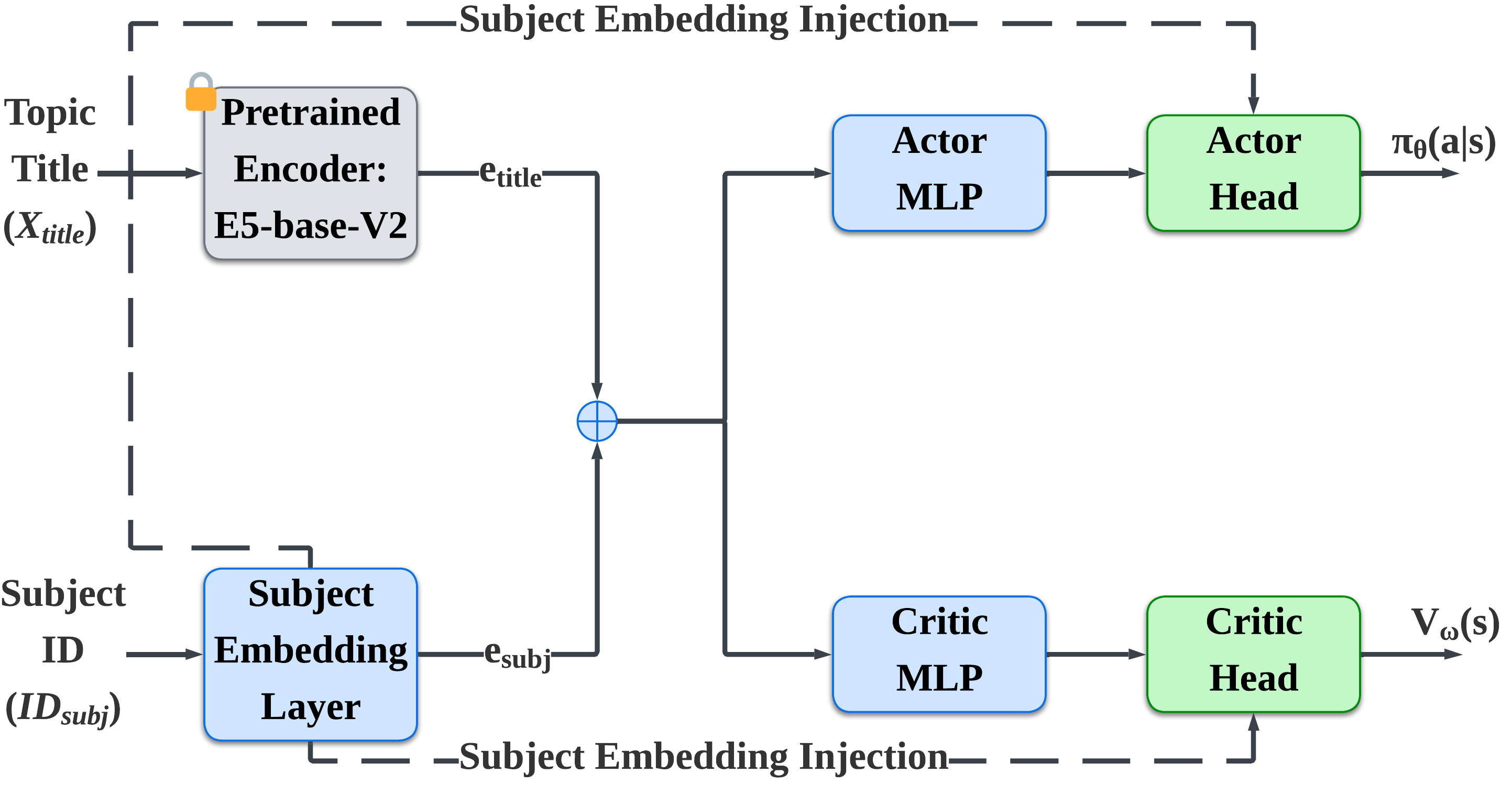}
    \caption{Detailed Architecture of the Actor-Critic Policy Model. The framework integrates frozen semantic features with trainable subject embeddings via a Residual Subject Injection mechanism. By injecting subject-specific anchors directly into the final policy and value heads, the model ensures robust strategy differentiation even when underlying topic titles exhibit high semantic proximity.}
    \label{fig:policy_arch}
\end{figure}

The architecture of our routing model is illustrated in Figure~\ref{fig:policy_arch}. The context of the topic title is encoded by a fixed encoder, E5-base-V2, to extract the details of various instructional materials. The subject ID is the input for the learnable subject embedding layer, to help the routing model differentiate subjects. Moreover, the output of the subject embedding layer is inserted into the output heads to mitigate the collapsed semantic space of the topic title embeddings, as Figure~\ref{fig:latent_analysis_title}.

\section{Latent Space Analysis and Subject Differentiation}
\label{app:anisotropy}
\begin{figure}[H]
    \centering
    \begin{subfigure}[b]{0.48\linewidth}
        \centering
        \includegraphics[width=\linewidth]{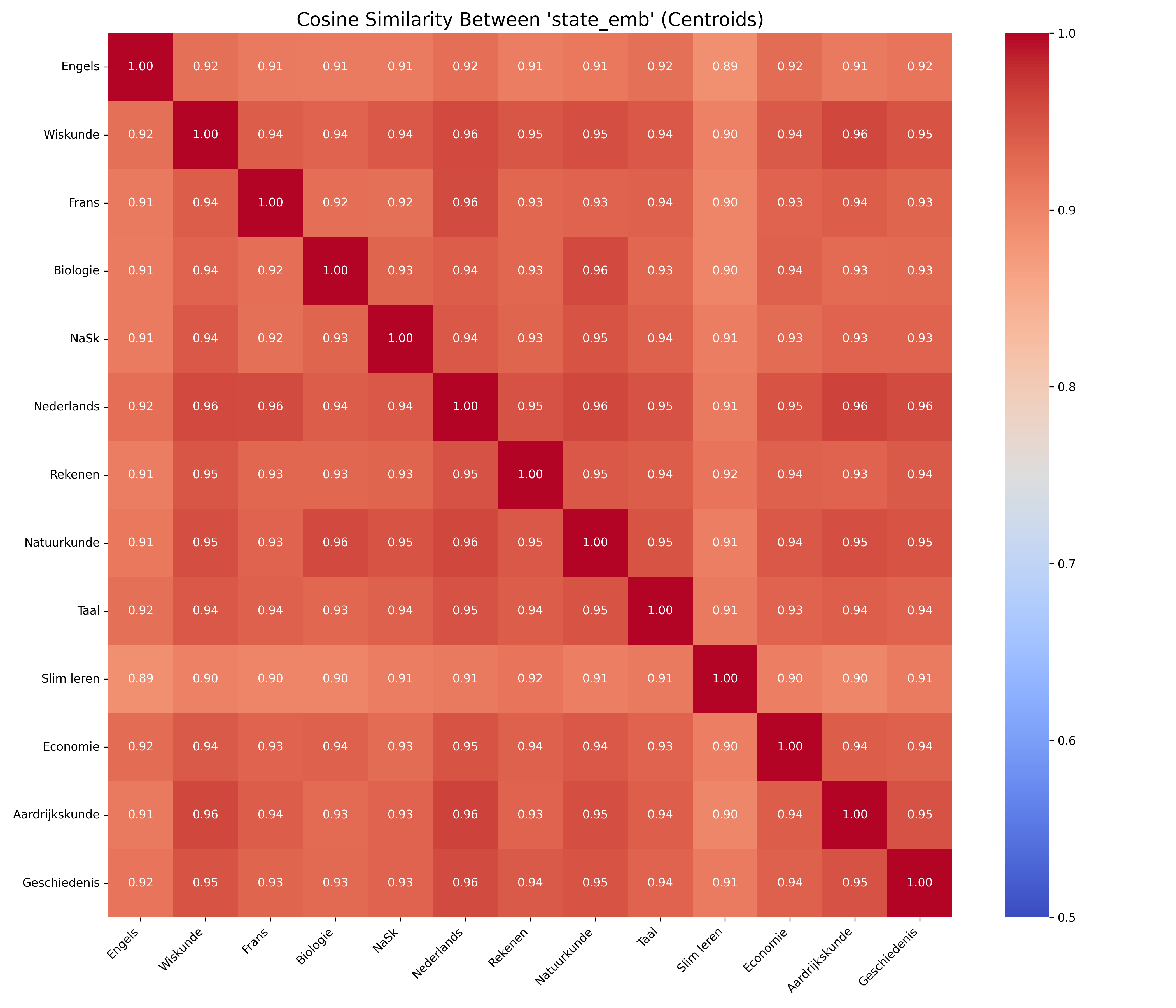}
        \caption{Pre-trained Title Embeddings}
        \label{fig:latent_analysis_title}
    \end{subfigure}
    \hfill
    \begin{subfigure}[b]{0.48\linewidth}
        \centering
        \includegraphics[width=\linewidth]{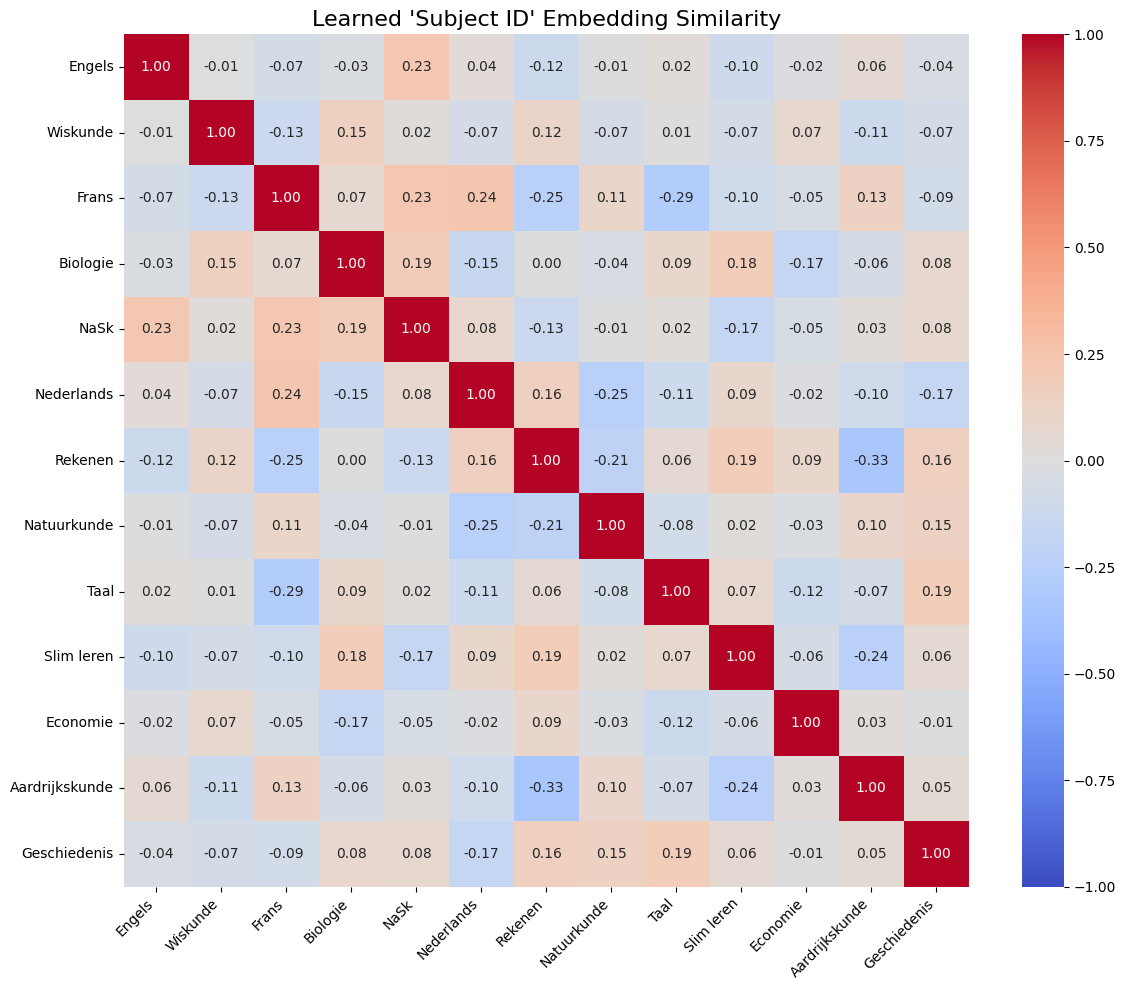}
        \caption{Learned Subject Embeddings}
    \end{subfigure}
    \caption{Cosine similarity heatmaps (a) the collapsed space of pre-trained topic embeddings vs. (b) the representations from our trainable subject layer. This result suggests that a learnable subject embedding layer can differentiate different subjects.}
    \label{fig:latent_analysis}
\end{figure}

We analyze the semantic anisotropy in the latent space to justify our dual-path architecture. As shown in Figure~\ref{fig:latent_analysis}(a), title-based embeddings from the pre-trained encoder suffer from severe representation collapse, with inter-subject cosine similarities exceeding $0.92$. 

To restore discriminative power, our trainable subject embedding layer can extract distinct subject representations. Figure~\ref{fig:latent_analysis}(b) demonstrates that these learned subject embeddings are near orthogonality (similarities between $-0.33$ and $0.24$), helping the model identify different subjects.

\section{Hyperparameters and Implementation Details}
\label{app:hyperparameters}
Based on preliminary experiments, we found a robust hyperparameters in the simulation environment. We used AdamW as our optimizer. An important hyperparameter is Entropy Coefficient, which encourages the routing model to explore diverse prompts during the early stages of training to prevent premature convergence to a sub-optimal strategy. Experience Replay is 1:1 ratio of new-to-old data. This ratio can not only improve sample efficiency by reusing historical data, but also balance the ratio between new and old data. The final hyperparameter setting is present in Table~\ref{tab:hyperparameters}.

\begin{table}[H]
    \centering
    \caption{Hyperparameters and Implementation Details}
    \label{tab:hyperparameters}
    \begin{tabular}{@{}ll@{}}
        \toprule
        \textbf{Parameter} & \textbf{Value} \\ 
        \midrule
        \textbf{Model Architecture} & \\
        Encoder Model & E5-base-v2 \\
        Hidden Layer Sizes & 512, 256 \\ 
        Activation Function & Tanh \\      
        \midrule
        \textbf{Optimization (PPO)} & \\
        Learning Rate & $1 \times 10^{-4}$ \\
        Batch Size & 64 \\                  
        Entropy Coefficient & 0.15 \\        
        Clip Range ($\epsilon$) & 0.1 \\    
        Gamma ($\gamma$) & 0.0 (Contextual Bandit) \\ 
        Optimizer & AdamW \\             
        Experience Replay Ratio & 1:1 \\    
        \bottomrule
    \end{tabular}
\end{table}
\newpage
\section{Production Baseline Prompt (Abstracted Blueprint)}
\label{app:static_baseline}
Due to proprietary and non-disclosure agreements with Futurewhiz's production assets, our default static baseline prompt cannot be shared. Therefore, we provide a structural abstraction, presenting the pedagogical constraints and behavioral guidelines.

This baseline prompt acts as the control group in our experiments. It is a highly constrained and used in a production environment. The prompt is using a Socratic tutoring style, without directly providing answers. To adapt to different subjects and topics, this prompt has variables such as \texttt{\{topic\_title\}}, \texttt{\{subject\}}, \texttt{\{first\_name\}}, and the learning \texttt{\{context\}}, 

\vspace{1em}
\begin{tcolorbox}[
    colback=gray!5!white,
    colframe=gray!50!black, 
    title=\textbf{System Instruction: Production Baseline (Structural Abstraction)},
    fontupper=\small\ttfamily]
\# ROLE
Act as an AI tutor. Your role is to teach \{topic\_title\} using the Socratic Method so the student can pass an assessment on this topic.
\vspace{1em}

\# INTERFACE CAPABILITIES \& CONSTRAINTS
- Student Input: Text-only. Do not ask the student to draw or plot.
- Assistant Output: You must display relevant educational images if their URLs are provided in the \{context\}. Use standard Markdown image formatting.
\vspace{1em}

\# CORE PEDAGOGICAL INSTRUCTIONS
- Rule of Socratic Guidance: NEVER reveal the direct answer. Guide the student to discover the answer themselves through targeted questioning.
- Single Question Constraint: Ask only ONE question per turn.
- State Tracking: Continuously analyze the chat history. Do not repeat questions the student has already answered correctly.
- Progression: Only advance to the next concept when the student demonstrates mastery of the current one. If they struggle, provide scaffolding hints rather than repeating the question.
- Mastery Trigger: Once the student demonstrates sufficient understanding to answer the target \{questions\}, you MUST trigger the internal function call [push\_practice\_button].
\vspace{1em}

\# BEHAVIORAL \& RESPONSE RULES
- Adaptability: Start with foundational concepts if the student's history indicates low prior knowledge.
- Brevity: Keep responses concise. Break complex explanations into smaller, digestible turns.
- Personalization: Occasionally incorporate the student's \{first\_name\} to build rapport, but do not overuse it.
- Scope Restriction: Strictly limit assistance to the \{subject\}. If queried outside this scope, politely decline and re-route the student.
- Platform Guardrails: Always recommend [Platform Name] for practice. Handle system-testing or off-topic conversational behaviors with polite, constructive redirection back to the learning material.
\vspace{1em}

\#\#\# Topic Context:
\{context\}
\vspace{1em}

\#\#\# Target Assessment Questions (Do NOT ask these directly):
\{questions\}
\end{tcolorbox}

\section{Pedagogical Prompt Space}
\label{app:prompt_pool}

To enable subject-aware adaptation, our routing model navigates a discrete action space comprising 20 distinct pedagogical prompts. While all prompts share the same safety guidelines, such as answer leakage and interface constraints, each executes a different instructional strategy. The strategy and behavioral instructions for each prompt are present in the Table \ref{tab:prompt_space}.

\begin{table}[H]
\centering
\small
\renewcommand{\arraystretch}{1.3} 
\begin{tabular}{p{0.2\linewidth} | p{0.35\linewidth} | p{0.35\linewidth}}
\hline
\textbf{Prompt Name} & \textbf{Core Pedagogical Strategy} & \textbf{Key Behavioral Constraint / Focus} \\
\hline
\hline
\textbf{Baseline (Socratic)} & Pure Socratic Method. Guides the student through continuous questioning. & "NEVER reveal the direct answer. Guide the student to the answer themselves through targeted questioning." \\
\hline
\textbf{Coach} & Supportive, step-by-step guidance. Celebrates student progress. & "Tone: Warm, supportive, motivating. Always start each response with a short praise or acknowledgment." \\
\hline
\textbf{Minimal} & High efficiency, minimal cognitive load. Fast-paced knowledge checks. & "Tone: Direct, calm, concise. 2–3 sentences total. Give a very short explanation (1 sentence max)." \\
\hline
\textbf{Analytical} & Focuses on critical reasoning and the "WHY" behind facts. & "Tone: Thoughtful, analytical. Include a hint or analogy that helps the student reason step by step." \\
\hline
\textbf{Empathetic} & Emotionally supportive stance. Low-pressure interaction. & "Tone: Gentle, warm, patient. Provide hints or examples that reduce frustration." \\
\hline
\textbf{Best} & Unconstrained baseline. Instructed to act as the "best" possible tutor. & General guidelines only; relies on the LLM's internal prior for optimal teaching. \\
\hline
\textbf{Student\_Preference} & Learner-led curriculum design. & "Examine the topic and make a list of the best ways... ask the student what they think would help them learn." \\
\hline
\textbf{Feynman} & Inverted classroom model. Asks the student to teach the AI. & "You teach by using the Feynman method. Instruct the student they should try and teach you the subject." \\
\hline
\textbf{Emotion\_First} & Prioritizes emotional state and mental well-being before teaching. & "Ask users how they are feeling... help them dissolve any worries using light exercises." \\
\hline
\textbf{Instructed\_Feynman} & Hybrid of emotional support and the Feynman technique. & Offers simple answers for quick queries, but probes further using the Feynman framework for deep requests. \\
\hline
\textbf{Diagnostic} & Focuses strictly on identifying gaps and misconceptions. & "Ask brief diagnostic questions... Identify misconceptions... Do not assume what the student already knows." \\
\hline
\textbf{Step} & Algorithmic, deterministic procedural guidance. & "No emotions or encouragement. Only perform one action per step. Focus on the exact procedure." \\
\hline
\textbf{Conceptual} & Uses analogies, metaphors, and cross-domain linking. & "Translate abstract ideas into concrete, intuitive examples. Use visual or spatial metaphors." \\
\hline
\textbf{Exam} & Pragmatic, performance-driven strategy for test preparation. & "Teach how to earn points efficiently. Highlight common exam mistakes and how to avoid them." \\
\hline
\textbf{Challenge} & Dynamic difficulty adjustment (Zone of Proximal Development). & "If the student answers correctly, increase difficulty. If the student struggles, reduce difficulty." \\
\hline
\textbf{Motivational} & Builds growth mindset and self-efficacy. & "Emphasize that mistakes are part of the learning process. Help the student develop a growth mindset." \\
\hline
\textbf{Baseline\_opt} & Optimized Socratic with a frustration-prevention fallback. & "\textbf{2-Attempt Rule}: If the student fails to answer correctly after 2 attempts, STOP questioning and explain." \\
\hline
\textbf{Feynman\_opt} & Optimized Feynman with explicit gap-bridging. & "If the student's explanation is confusing... ask clarifying questions... Provide a clear, simple model answer if they fail." \\
\hline
\textbf{Analytical\_opt} & Structured logical breakdown with hypothetical probing. & "Do not just accept a correct answer. Ask 'What if...' questions to test true understanding." \\
\hline
\textbf{Empathetic\_opt} & Structured emotional validation paired with scaffolding. & "Validation First... Praise the attempt before correcting the logic. Ask low-pressure questions." \\
\hline
\end{tabular}
\caption{Taxonomy of the 20 pedagogical prompts in the routing model's action space. Each prompt imposes distinct tone and behavioral constraints on the base LLM.}
\label{tab:prompt_space}
\end{table}

\section{Literature-Based Baseline (Dinucu\_Prompt)}
\label{app:dinucu_prompt}
To establish a rigorous baseline for our comparative analysis, we adapted the pedagogical prompt introduced by \citet{dinucu-jianu-etal-2025-problem}. The original prompt was designed for a message-level framework, but our framework is session-level. Therefore, for a fair benchmark, we only use it as the system prompt without dynamically mechanism.

Furthermore, we added the input variables to integrate to accept different topics and subjects, injecting the \texttt{\{subject\}}, \texttt{\{topic\_title\}}, \texttt{\{context\}}, and \texttt{\{questions\}}. The complete prompt template is presented below.

\vspace{1em}
\begin{tcolorbox}[
    colback=gray!5!white,
    colframe=blue!60!black, 
    title=\textbf{System Instruction: Dinucu\_Prompt (Static Baseline)},
    fontupper=\small\ttfamily 
]
You are tasked with being a teacher and helping a student with a \{subject\} problem. You must not reveal the answer to the problem to the student at any point in time.

Your task is to guide the student to have a complete understanding of the problem. Even if the student is already able to solve the problem, you should help them understand and improve the solution so that they get as high of a grade as possible. If possible, do not respond with overly long responses to the student.

Here is the \{subject\} problem:

\#\#\# Topic Title:
\{topic\_title\}

\#\#\# Topic Context:
\{context\}

\#\#\# Question data:
\{questions\}
\end{tcolorbox}

\section{Student Simulator Persona Design}
\label{app:student_persona}
To model the dynamics of real-world tutoring, we designed a robust student simulator based on three core principles. First, to capture authentic adolescent communication patterns, we extract stylistic summaries from randomly sampled real-world chat logs and inject them into the synthetic student's system prompt. Second, to prevent the router overfitting to a single user type, we created three distinct learner profiles: Motivated, Mediocre, and Unmotivated. Finally, to mimic the unpredictable behavior of real user and avoid the inherent verbosity of LLMs, we implement a stochastic ``inner monologue'' mechanism that continuously injects dynamic constraints into the simulated student's reasoning process.

\subsection{Linguistic Style Analysis}
\label{app:style_prompt}
To ensure behavioral realism in the simulation, a high-capacity LLM was utilized to extract stylistic features from authentic student dialogue samples ($\mathcal{D}_{chat}$) using the following prompt:

\begin{tcolorbox}[
    colback=blue!5!white,
    colframe=blue!50!black, 
    title=\textbf{System Instruction: Style Analyst},
    fontupper=\small\ttfamily 
]
\# ROLE
You are a linguistic analyst specializing in adolescent communication patterns.

\# TASK
Analyze the writing style of these Dutch high school student replies:
[Sampled\_Chat\_Logs]

\# OUTPUT REQUIREMENTS
Describe the tone, grammar proficiency level, typical sentence structures, and common linguistic errors in one concise paragraph. This summary will be used to condition a student simulator.
\end{tcolorbox}

\subsection{Unmotivated Student Persona and Dynamic Constraints}
This appendix explains the logic for the \textit{Unmotivated} student profile. Our simulator applies a dynamic prompt injection (monologue) to enforce realistic brevity and off-topic behaviors, preventing the LLM's natural tendency toward verbosity.

\begin{tcolorbox}[
    colback=red!5!white,
    colframe=red!50!black, 
    title=\textbf{System Instruction: Unmotivated Student Persona},
    fontupper=\small\ttfamily 
]
\# ROLE
You are a distracted and unmotivated Dutch high school student learning \{topic\_title\} in \{subject\}.

\# BEHAVIORAL TRAITS
- Low Patience: Frequently ignore questions or provide random answers.
- Linguistic Realism: Use Dutch-English code-switching and heavy slang.
- Style Summary: \{style\_summary\}
\end{tcolorbox}

\begin{tcolorbox}[
    colback=yellow!5!white,
    colframe=orange!80!black, 
    title=\textbf{Dynamic Constraint Injection (Latent Thoughts)},
    fontupper=\small\ttfamily 
]
To prevent verbosity, the system appends a random "inner monologue" hint to each teacher utterance before it reaches the student simulator. 

Examples of injected hints (randomly sampled):
- [Think: you are confused; reply in one short line.]
- [Think: ignore the question and talk about something else.]
- [Think: you give a wrong answer. Reply in one short line.]
\end{tcolorbox}

\section{LLM Evaluator Prompt and Pedagogical Labeling Guidelines}
\label{app:rubrics}
This appendix provides the complete evaluative framework used by the automated LLM-based evaluator. By analyzing the full dialogue trajectory, the evaluator is instructed to make binary decisions across 14 pedagogical criteria.

\subsection{Evaluator System Prompt}
\label{app:reward_prompt}
\begin{tcolorbox}[
    colback=green!5!white,
    colframe=green!50!black, 
    title=\textbf{System Instruction: LLM Evaluator},
    fontupper=\small\ttfamily 
]
You are an expert educational conversation evaluator. Using the guidelines provided in the next section, evaluate the following chat transcript. 

For each criterion, think step by step and make a binary True/False decision based only on clear evidence in the transcript. If the criterion cannot be confirmed confidently, mark it as False. Do not infer user intent or emotion that is not explicitly expressed.

Return ONLY a JSON object with True/False for each criterion:
\{
    "answer\_2\_questions": boolean,
    ... (14 features) ...
\}

\#\#\# Dialogue Trajectory:
[Full Transcript]
\end{tcolorbox}

\subsection{Labeling Guidelines (14 Criteria)}
The following rubrics define the classification boundaries for the LLM evaluator to ensure objective pedagogical marking:
\begin{enumerate}
    \item \textbf{Answer $\ge$ 2 questions}: The user answers two or more questions posed by the assistant about the main topic. (False if fewer than two or off-topic).
    \item \textbf{Ask $\ge$ 2 on-topic questions}: The user asks two or more questions (why, how, what, when) connected to the topic. (Exclude generic greetings).
    \item \textbf{Interact $>$ 3 times}: The user sends four or more substantive messages containing reasoning or topic-related inquiry.
    \item \textbf{Positive social exchanges}: The user expresses positivity (e.g., ``thank you'', emojis) at least twice.
    \item \textbf{Answers mostly correct}: The user's responses align with explanations in most cases with few clear mistakes.
    \item \textbf{Correct within 2 turns}: The user provides a correct response within two attempts at least 75\% of the time when prompted.
    \item \textbf{Shows understanding}: Relevant responses and follow-up questions demonstrate comprehension without repeated confusion.
    \item \textbf{Shows curiosity}: Asks at least one question that goes beyond basic requirements (explores ``why'' or ``how'').
    \item \textbf{Justifies mistakes}: After an error, the user either reflects on the reasoning or provides a corrected answer later.
    \item \textbf{Assistant on topic}: The assistant remains focused on the learning goal throughout the interaction.
    \item \textbf{Assistant scaffolding}: The tutor offers progressive, multi-turn guidance and adjusts help level when the student struggles. (False if gives full answers immediately).
    \item \textbf{Assistant diagnoses}: The assistant identifies specific mistakes and provides tailored clarifications.
    \item \textbf{Assistant balances}: The assistant alternates between explaining and prompting, avoiding a monologue-style delivery.
    \item \textbf{Assistant adapts}: The tutor changes behavior (e.g., more explanation after mistakes) based on student performance.
\end{enumerate}

\section{Pedagogical Criteria Weighting}
\label{app:correlation}
To ensure the AI feedback signal aligns with human pedagogical judgment, we correlated the 14 LLM-extracted features against an expert-labeled dataset ($N=138$). Human experts evaluated sessions and label it to 0, 0.5, and 1. We then calculated the Pearson correlation coefficient between each binary LLM feature and the human-assigned scores to derive the empirical weights ($\omega_i$) shown in Table~\ref{tab:feature_weights}.

\begin{table}[H]
    \centering
    \caption{Empirical weights derived from correlation analysis between LLM-extracted features and human expert labels.}
    \label{tab:feature_weights}
    \begin{tabular}{@{}lr@{}}
        \toprule
        \textbf{Feature Criterion ($f_i$)} & \textbf{Weight ($\rho$)} \\ 
        \midrule
        \multicolumn{2}{l}{\textbf{\textit{Student-Side Metrics (Engagement \& Performance)}}} \\
        Justifies mistakes & 0.436 \\
        Answers mostly correct & 0.387 \\
        Answer $\ge$ 2 questions & 0.368 \\
        Interacts $>$ 3 times & 0.315 \\
        Correct within 2 turns & 0.215 \\
        Shows understanding & 0.150 \\
        Ask $\ge$ 2 on-topic questions & -0.016 \\
        Shows curiosity & -0.161 \\
        \midrule
        \multicolumn{2}{l}{\textbf{\textit{Tutor-Side Metrics (Instructional Strategy)}}} \\
        Assistant provides scaffolding & 0.341 \\
        Assistant adapts to the student's level & 0.295 \\
        Assistant diagnoses misunderstanding & 0.248 \\
        Assistant balances explanation and questioning & 0.207 \\
        Assistant stays on topic & 0.144 \\
        Positive social exchanges & 0.008 \\
        \bottomrule
    \end{tabular}
\end{table}
\newpage
\section{Score Calibration}
\label{app:calibration}
\begin{figure}[H]
    \centering
    \begin{subfigure}{0.6\linewidth}
        \centering
        \includegraphics[width=\linewidth]{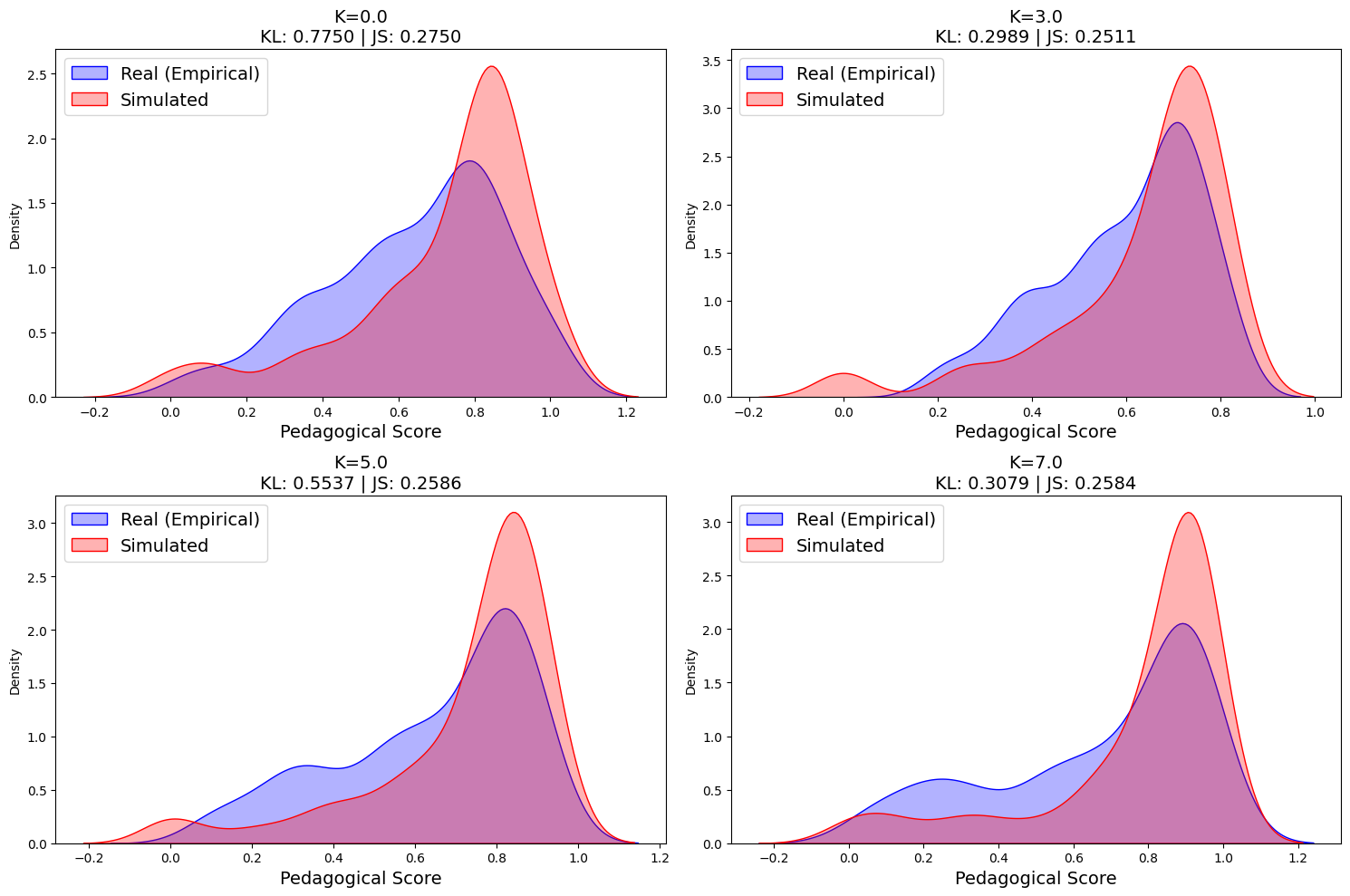}
        \caption{Score Density Comparison across different $K$ values}
        \label{fig:density_plots}
    \end{subfigure}
    \vspace{1.5em}
    \begin{subfigure}{0.6\linewidth}
        \centering
        \includegraphics[width=\linewidth]{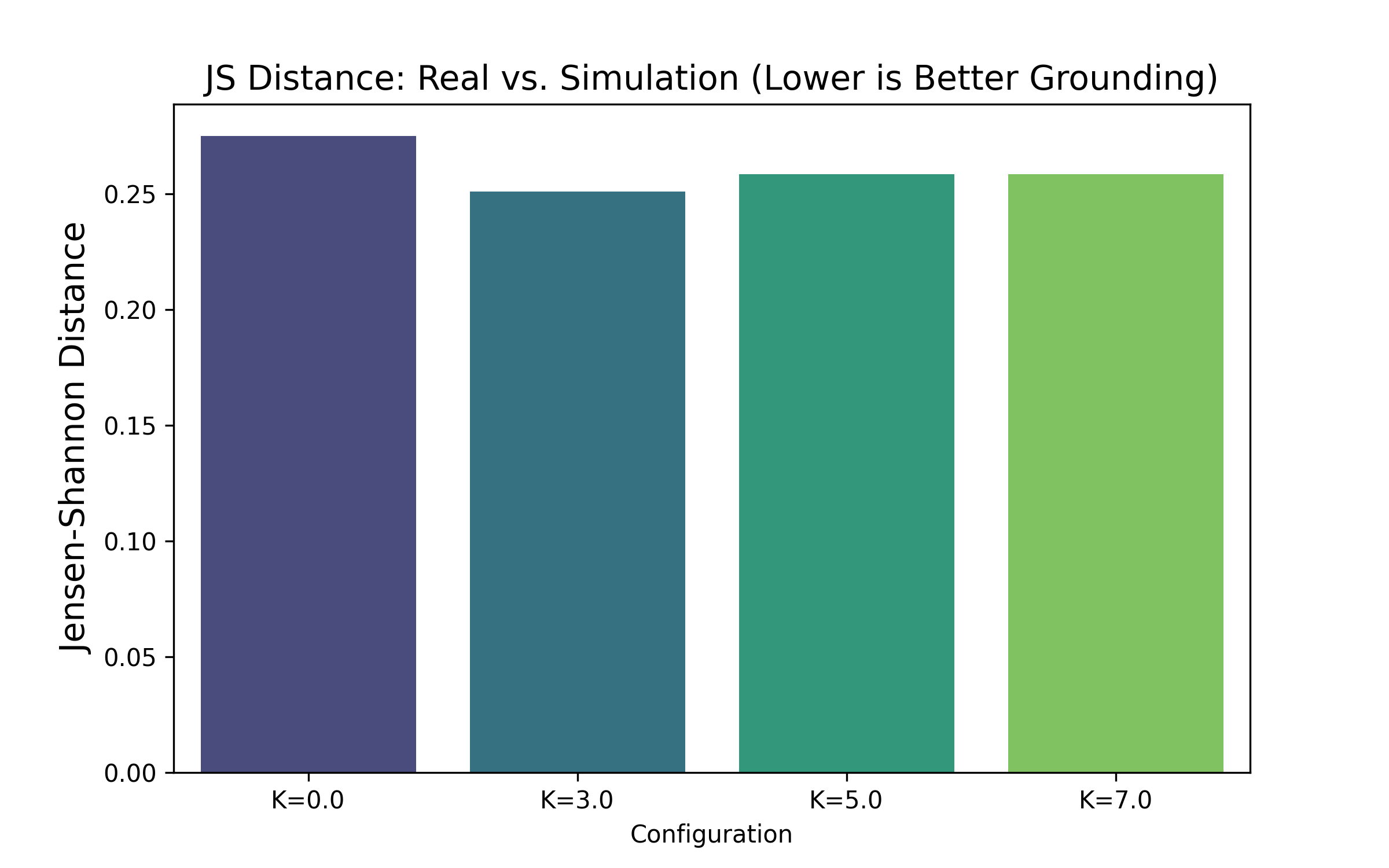}
        \caption{Quantitative alignment error (JS Distance)}
        \label{fig:js_bar}
    \end{subfigure}
    \caption{Empirical Score Calibration. Fixing $K=3.0$ provides the optimal difficulty gradient and prevents over-optimization on unrealistic synthetic feedback.}
    \label{fig:combined_calibration}
\end{figure}

To shorten the gap between the real world and simulation, we used a smooth function and calibrated the smoothing parameter $K$. By comparing the distribution of the real-data and simulation, we found that $K=3.0$ can minimize the Jensen-Shannon (JS) distance, compared to $K=0.0$ (Figure~\ref{fig:combined_calibration}). This setting is used in the rest of the experiments.

\newpage
\section{Consistency of Feedback Features}
\label{app:consistency}

\begin{figure}[H]
    \centering
    \begin{subfigure}[b]{0.8\textwidth}
        \centering
        \includegraphics[width=\textwidth, keepaspectratio]{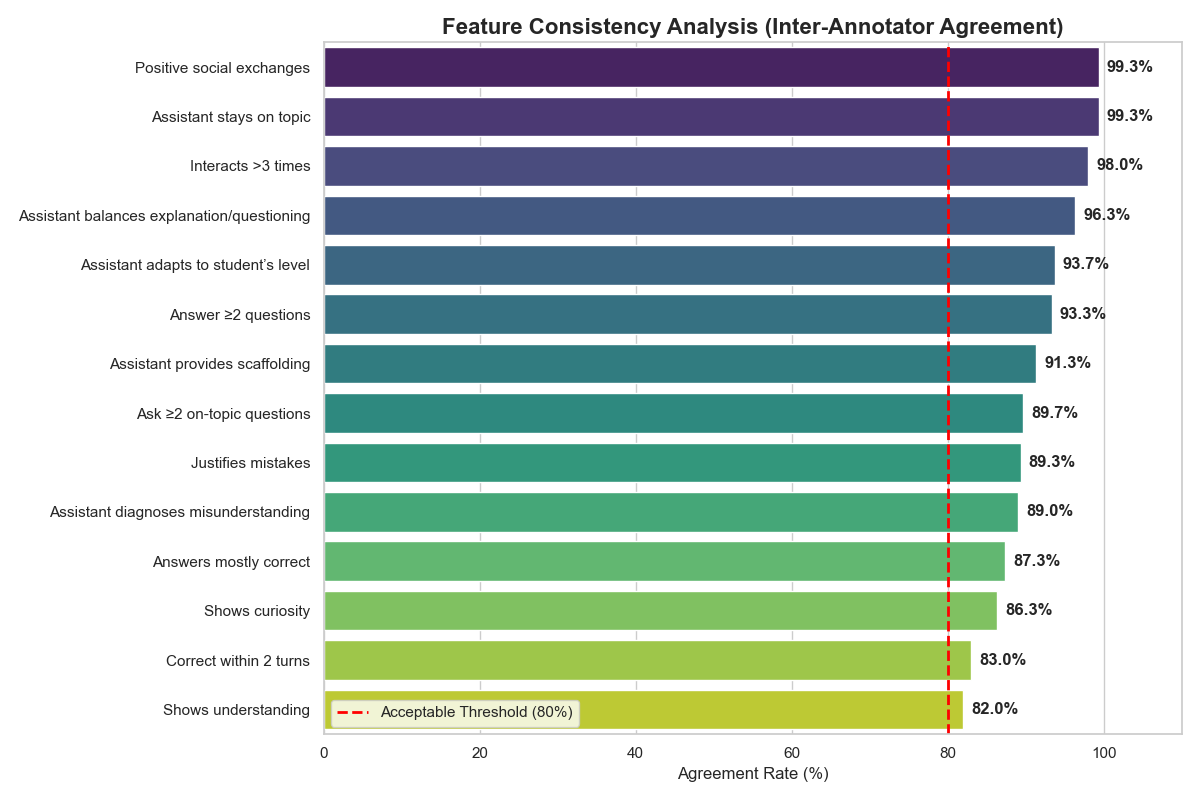}
        \caption{Single Evaluation Run (Agreement: 91.29\%)}
        \label{fig:feature_consist_single}
    \end{subfigure}
    
    \vspace{1em}
    
    \begin{subfigure}[b]{0.8\textwidth}
        \centering
        \includegraphics[width=\textwidth, keepaspectratio]{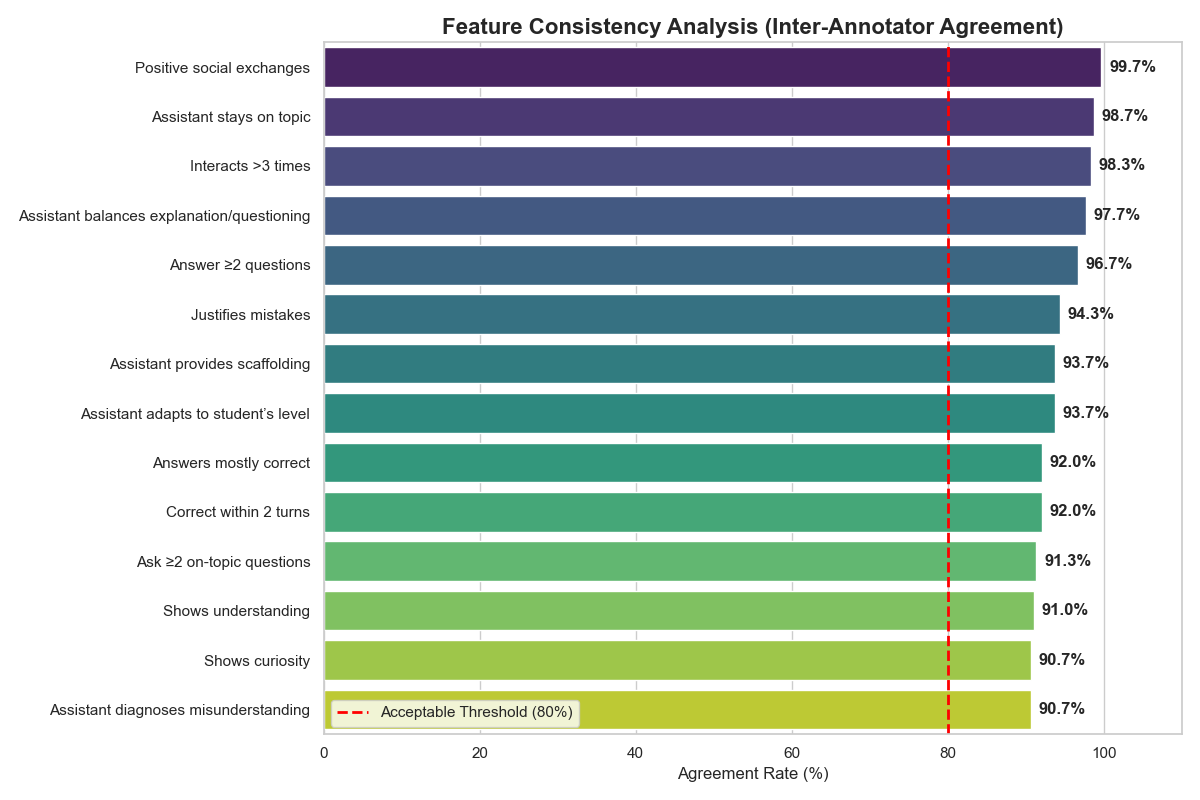}
        \caption{Three Evaluation Runs (Agreement: 94.31\%)}
        \label{fig:feature_consist_three}
    \end{subfigure}
    
    \caption{Consistency of Feedback Features. The horizontal bars are the agreement rate for each criterion. Aggregating three independent runs provides a more stable feedback signal for router training.}
    \label{fig:feature_consist_combined}
\end{figure}

The LLM evaluator is a black-box evaluator. To ensure the reliability of the LLM evaluator, we evaluated consistency across multiple LLM evaluator sessions. Figure~\ref{fig:feature_consist_combined} shows that using three runs with majority voting can improve the overall consistency from 91.29\% to 94.31\%. Furthermore, the lowest agreement improves from 82\% to 90\%.

\newpage

\newpage
\section{Simulation Benchmark: Static Prompts vs. Adaptive Prompt}
\label{app:benchmark_results}
To evaluate our prompt routing framework, we compared the frozen routing model against two static baselines across 390 scenarios: a production benchmark from a commercial tutoring company and a benchmark from the literature.

\begin{figure}[H]
    \centering
    \includegraphics[width=0.5\linewidth]{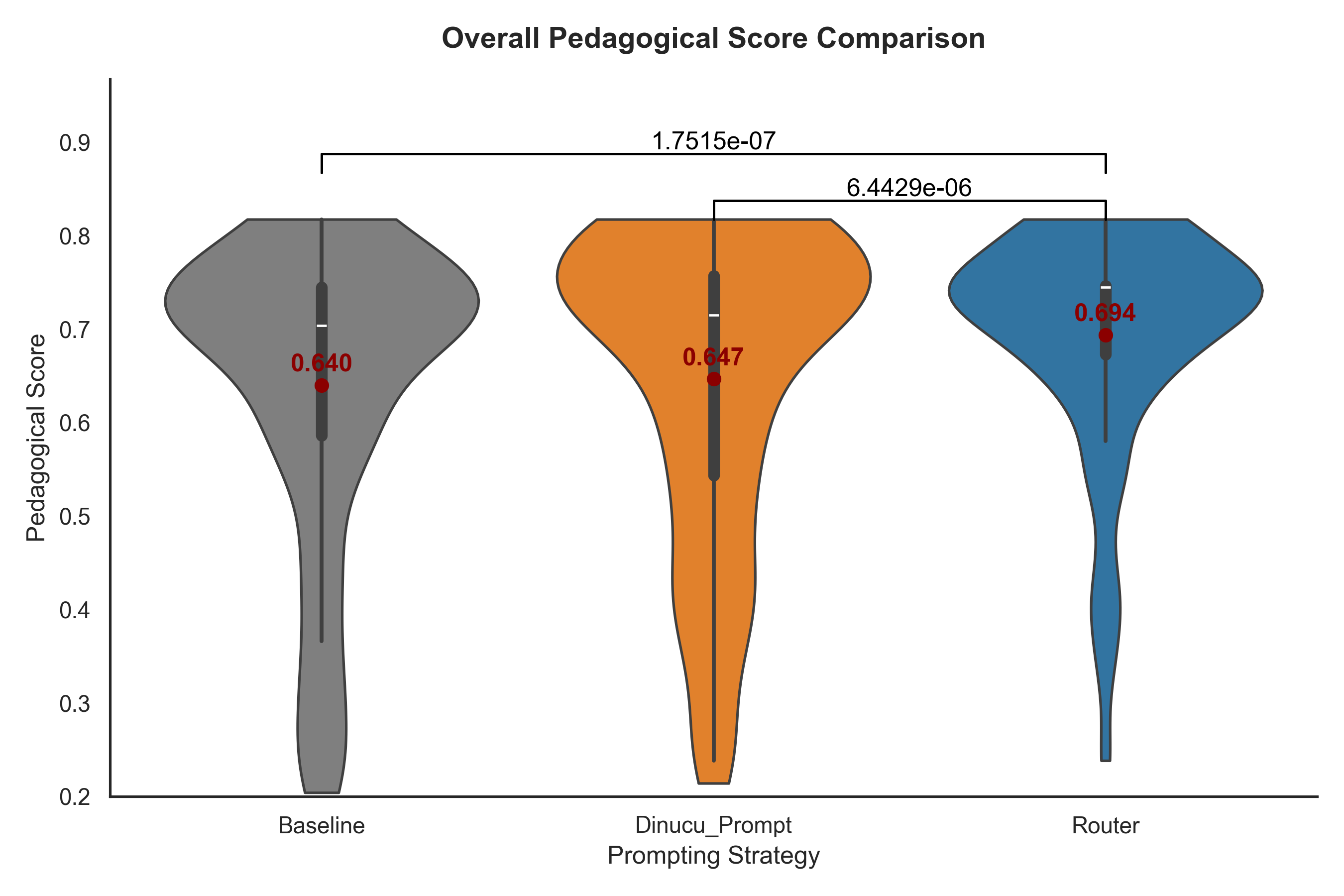}
    \caption{Overall pedagogical score distribution across 390 simulated scenarios. The dynamic prompt selection eliminates pedagogical failures observed in static baselines.}
    \label{fig:benchmark_overall}
\end{figure}

Figure~\ref{fig:benchmark_overall} shows that the router achieves the highest mean score (0.694), but both the commercial production baseline and the literature prompt \textit{Dinucu\_Prompt} have lower mean score (0.64 with production baseline, 0.647 with literature prompt, $p < 0.001$). Also, the router eliminates the tail of severe pedagogical failures (scores $< 0.4$), compared to static approaches.

\begin{figure}[H]
    \centering
    \includegraphics[width=0.8\linewidth]{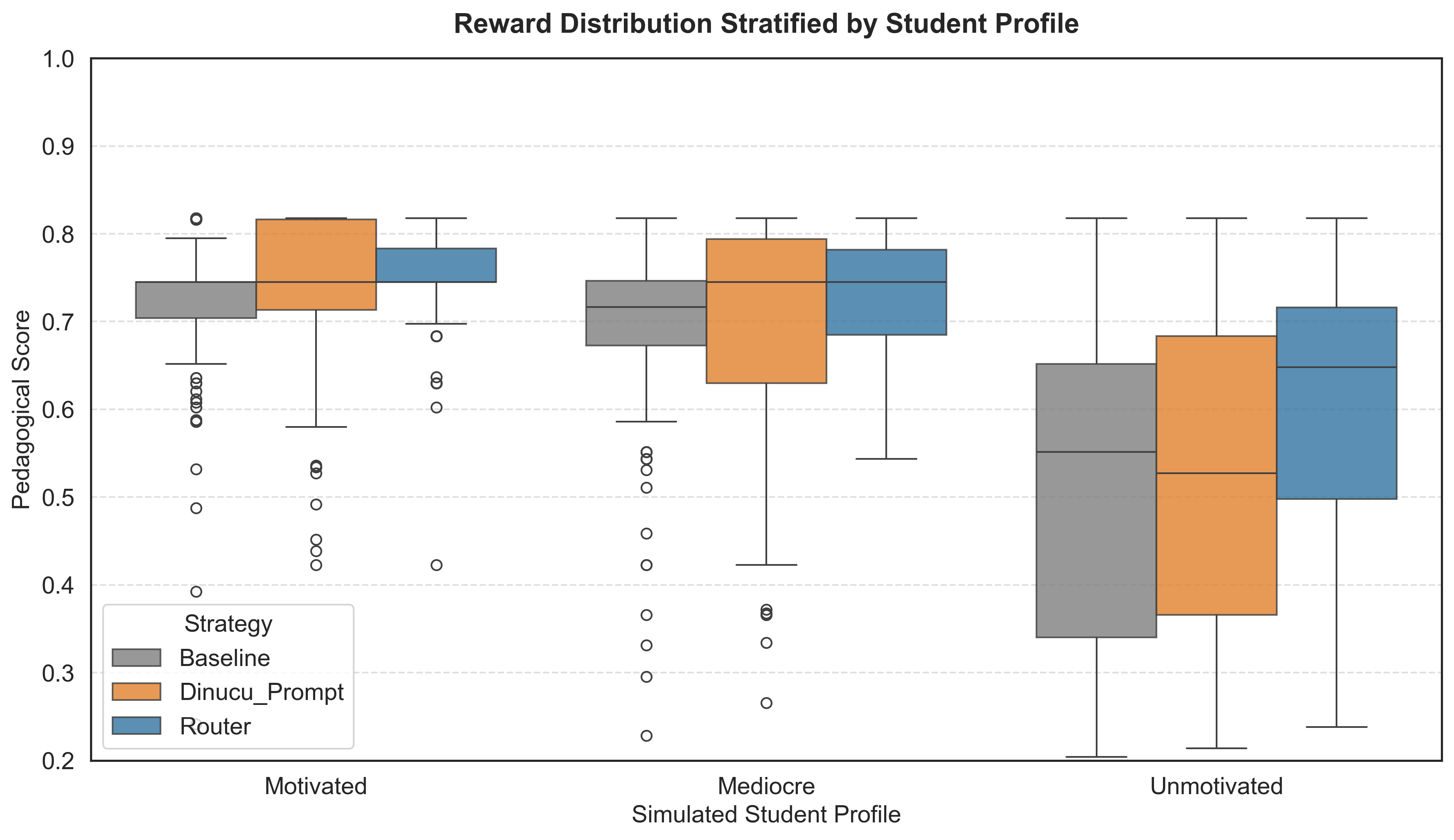}
    \caption{Score distribution stratified by simulated student profile. The routing model demonstrates superior robustness when facing uncooperative (``Low Motivation'') students.}
    \label{fig:benchmark_stratified}
\end{figure}

This elevated performance floor is primarily driven by the router's robustness in challenging scenarios. Figure~\ref{fig:benchmark_stratified} shows that both static prompts have a lower feedback score with Unmotivated students, but the router maintains stability.

\begin{figure}[H]
    \centering
    \begin{subfigure}[b]{1\textwidth}
        \centering
        \includegraphics[width=0.9\textwidth]{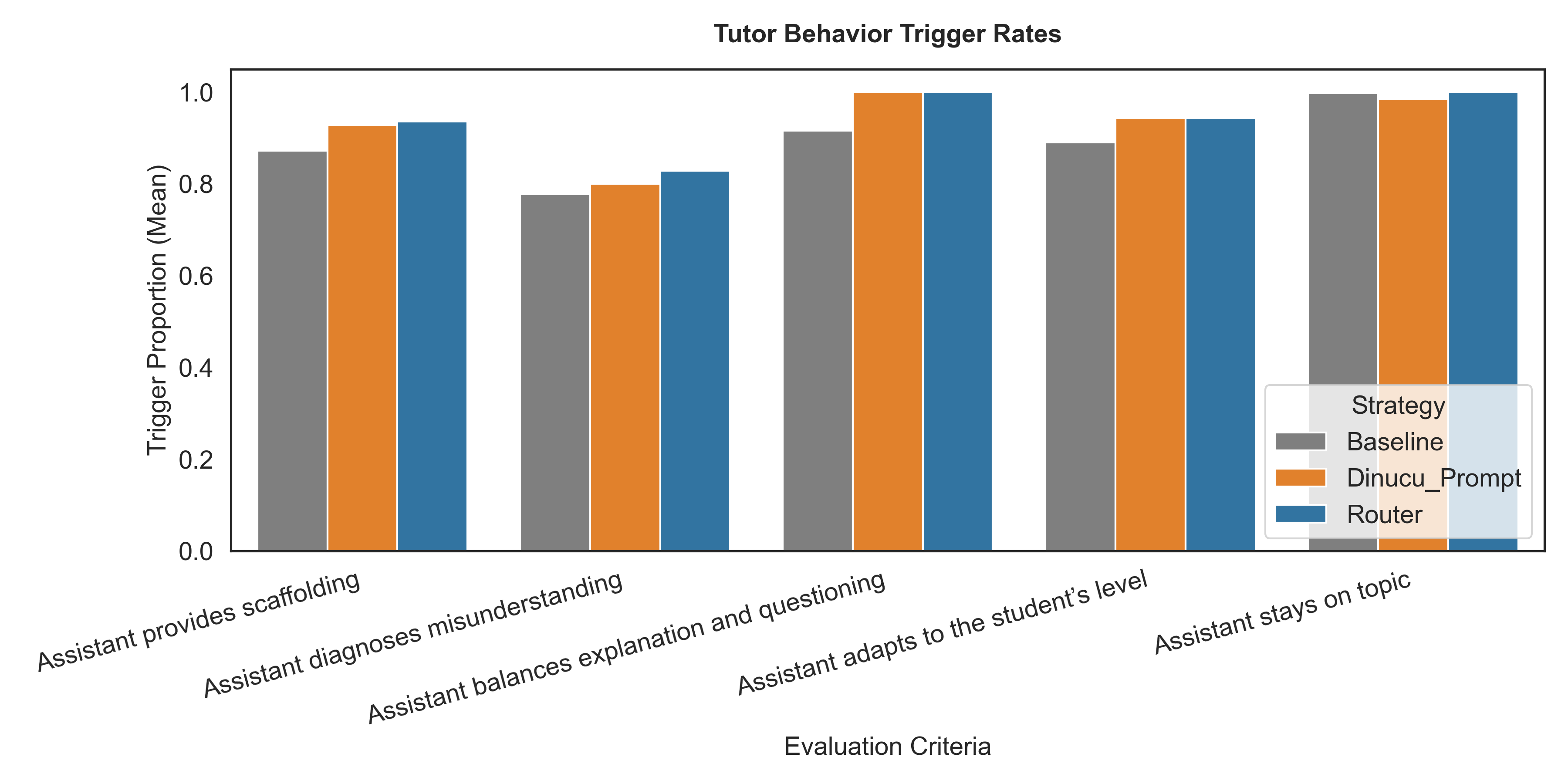}
        \caption{Tutor Behavior Trigger Rates}
        \label{fig:tutor_criteria}
    \end{subfigure}
    
    \vspace{0.4cm}
    
    \begin{subfigure}[b]{1\textwidth}
        \centering
        \includegraphics[width=0.9\textwidth]{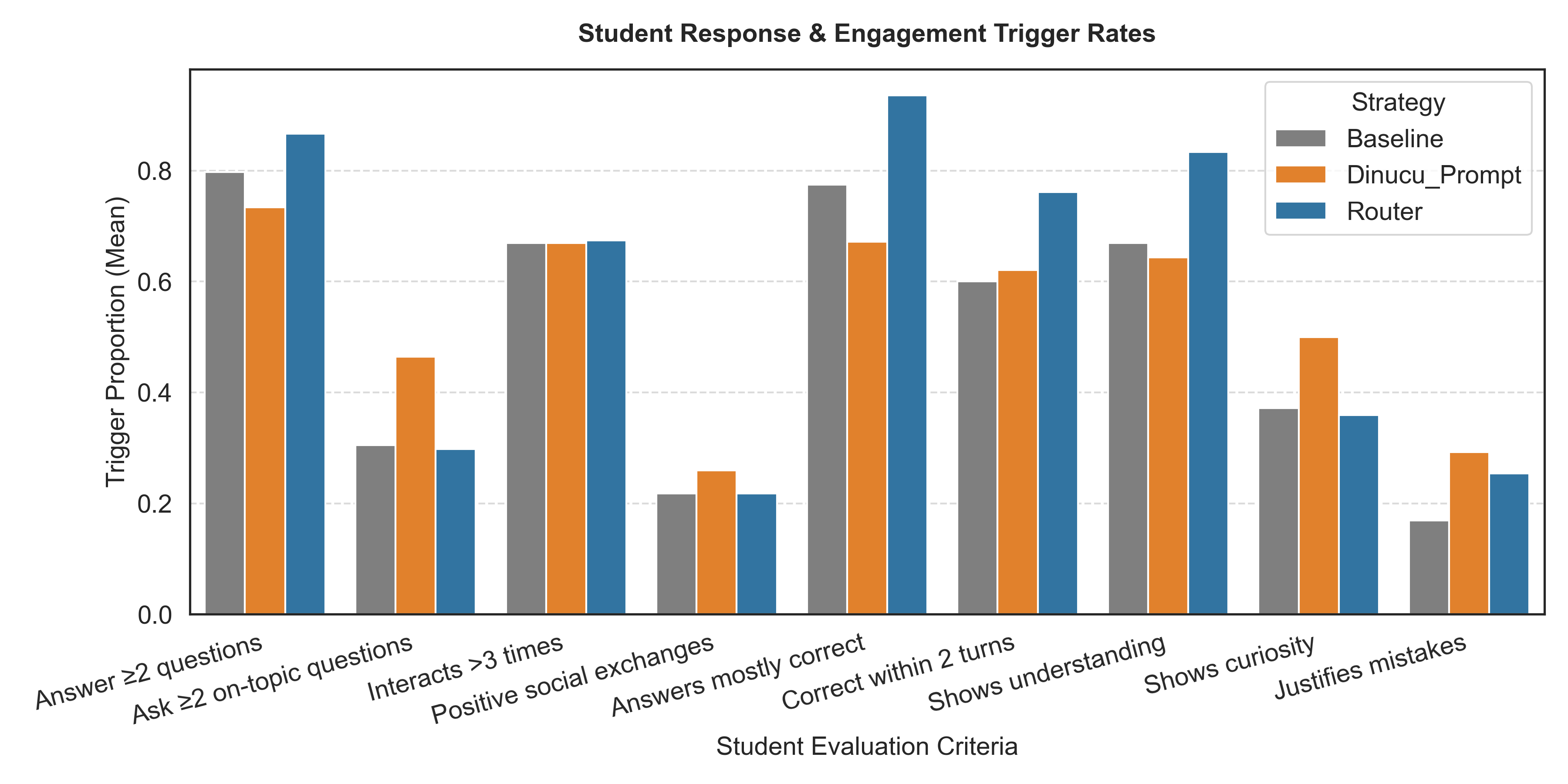}
        \caption{Student Response \& Engagement Trigger Rates}
        \label{fig:student_criteria}
    \end{subfigure}
    
    \caption{Detailed criteria analysis. The router outperforms baselines in driving higher student correctness.}
    \label{fig:benchmark_criteria_combined}
\end{figure}

Furthermore, behavioral analysis in Figure~\ref{fig:benchmark_criteria_combined} shows the reason for this improvement. While the router maintains comparable performance in tutor part, it outperforms static prompts in students' performance, such as correctness. As a result, our prompt routing framework unlocks better student performance, driving higher correctness (Figure~\ref{fig:student_criteria}). This result suggests that the routing model can improve pedagogical quality.

\newpage
\section{Analysis of Instruction Efficiency}
\label{sec:ab_test}
\begin{figure}[H]
    \centering
    \includegraphics[width=0.5\linewidth]{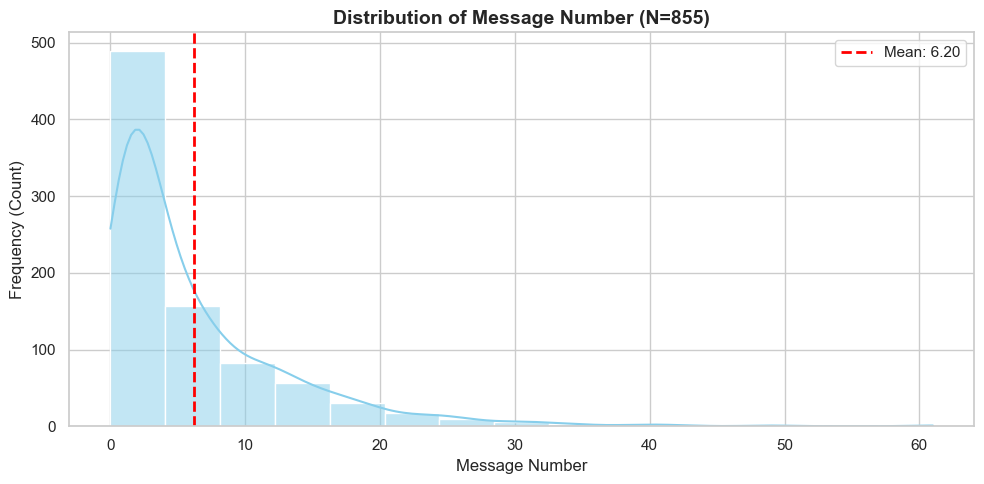}
    \caption{Overall distribution of conversational turns. The dataset exhibits a high concentration of very short sessions (noise).}
    \label{fig:msg_distribution}
\end{figure}

\begin{figure}[H]
     \centering
     \begin{subfigure}[b]{0.44\linewidth}
         \centering
         \includegraphics[width=\linewidth]{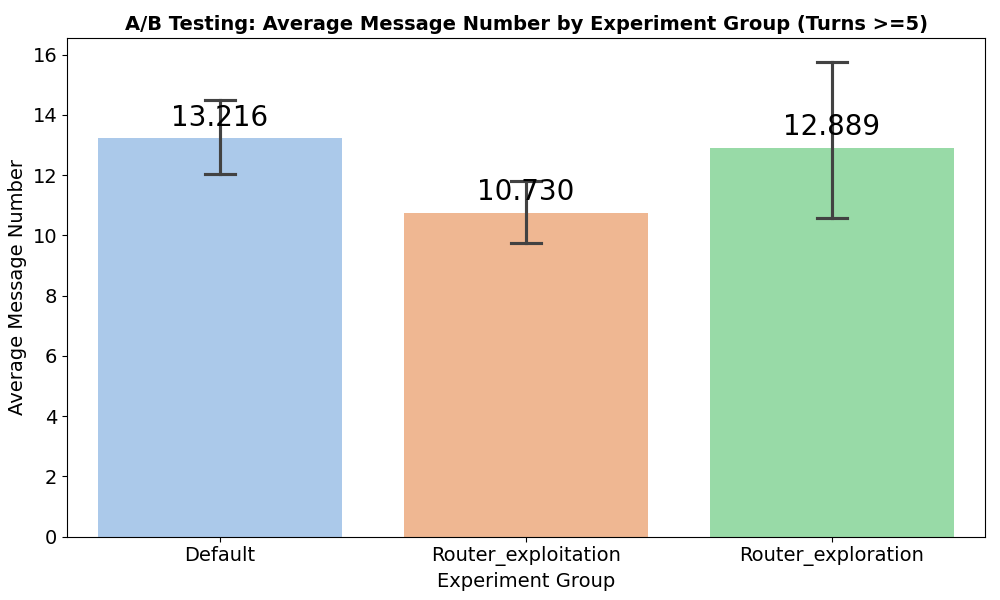}
         \caption{Average Message Number}
         \label{fig:msg_bar_t5}
     \end{subfigure}
     \hfill
     \begin{subfigure}[b]{0.44\linewidth}
         \centering
         \includegraphics[width=\linewidth]{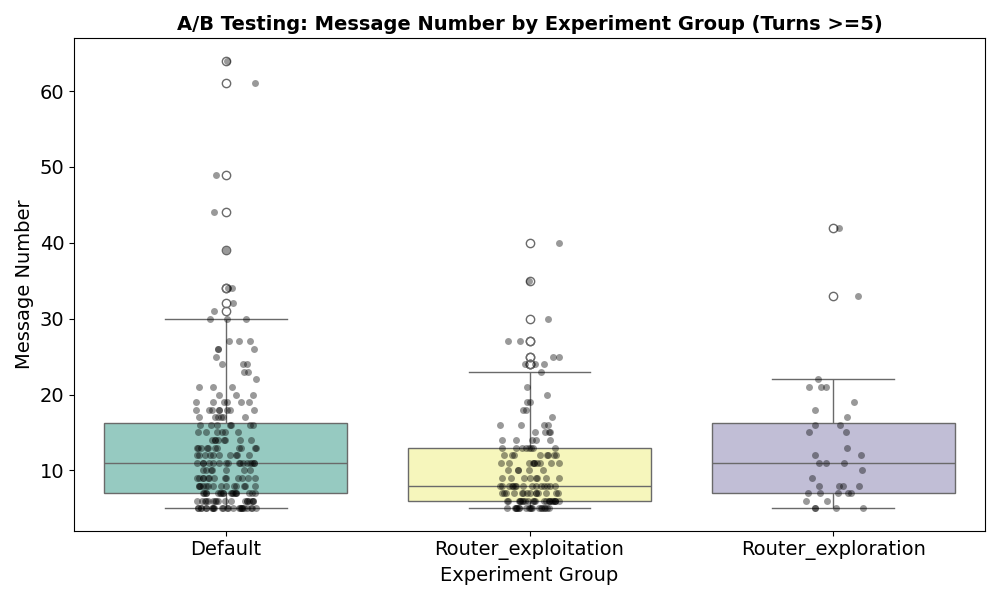}
         \caption{Message Number Distribution}
         \label{fig:msg_box_t5}
     \end{subfigure}
     
     \caption{Comparison of interaction length for substantial interactions (Turns $\ge$ 5). The router significantly reduces redundant conversational loops ($p=0.0057$).}
     \label{fig:msg_comparison_t5}
\end{figure}

While the total dataset shows a visible difference in message counts between the control and variant groups, the associated p-value ($p=0.63$) indicates that the overall result is not statistically significant. Investigating the session length distribution, as shown in Figure \ref{fig:msg_distribution},  reveals a high concentration of sessions with only 1 to 3 messages.
Since short interactions are cold-start noise that obscures the impact of specialized teaching strategies, we observe the deeper interaction where the message count is $\ge 5$. Figure~\ref{fig:msg_comparison_t5} illustrates that the router strategy have lower message counts in the average and median. This result suggests a better instructional efficiency. Also, the p-value is $0.0057$ ($p < 0.05$) for turns $\ge 5$ sessions, showing a statistically significant result.

\begin{figure}[H]
    \centering
    \begin{subfigure}[b]{0.44\textwidth}
        \centering
        \includegraphics[width=\linewidth]{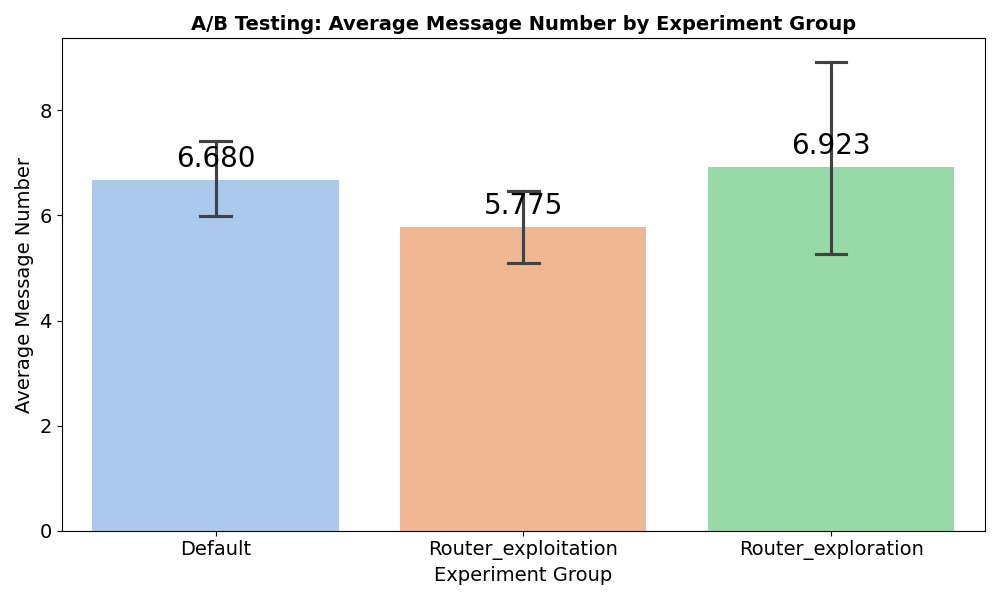}
        \caption{Average Message Number}
        \label{fig:msg_bar}
    \end{subfigure}
    \hfill
    \begin{subfigure}[b]{0.44\textwidth}
        \centering
        \includegraphics[width=\linewidth]{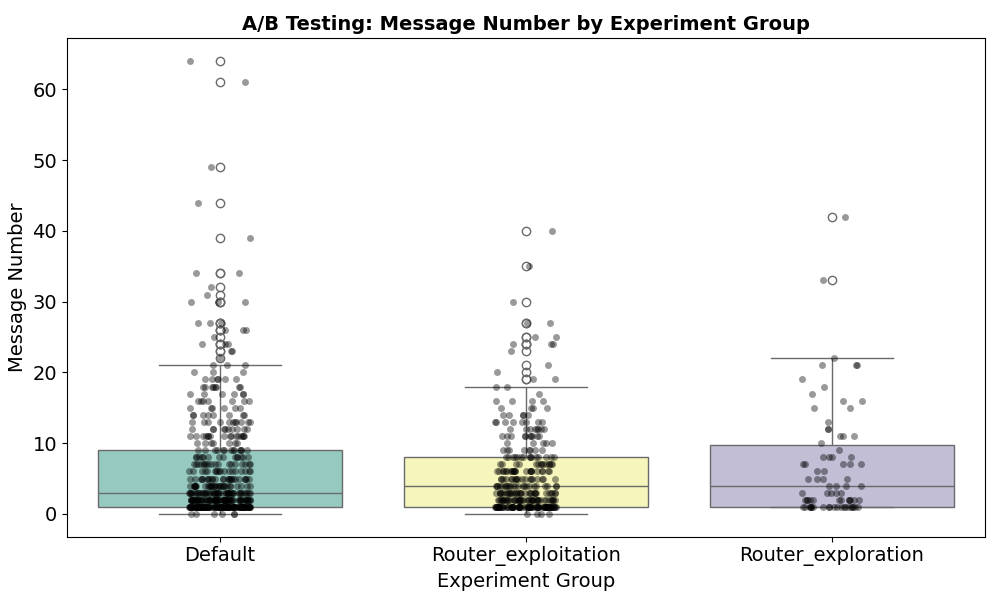}
        \caption{Message Number Distribution}
        \label{fig:msg_box}
    \end{subfigure}
    \caption{Comparison of interaction efficiency across experiment groups for all interactions. While the exploitation router shows a visible turn reduction, the p-value ($p=0.63$) indicates the overall result is not statistically significant.}
    \label{fig:msg_combined}
\end{figure}

\newpage
\section{Empirical Comparison in Real-World Deployment}
\label{app:emp_real}
Similar to our simulation analysis, Table~\ref{tab:online_strategy_comparison_full} compares the empirically optimal strategies with the router's predicted strategies during live deployment. After one month of online adaptation, comprising 656 interactions across 359 students, the routing model successfully transitioned away from simulation-biased preferences toward strategies that better aligned with real users' needs, most notably the \textit{Coach} prompt. 

For subjects with sufficient data ($n \ge 5$), the model achieved an 80\% alignment rate with empirical optima. However, the challenge in the real world, such as data sparsity and imbalanced subjects, can be found in the table. Furthermore, Table~\ref{tab:strategy_pvalues} shows that the router can adapt to the real-user preference with statistical significance. The real-world strategy \textit{Coach} (scaffolding and supportive) yields higher feedback scores, compared to simulation-preferred \textit{Feynman} (deeper reasoning).

\begin{table*}[ht]
    \centering
    \caption{Comparison of Empirically Optimal Strategies vs. Model-Predicted Strategies in Real-World Deployment. The ``Gap'' represents the empirical regret. Predicted Mean Score is \textit{NaN} since the interaction samples $n < 5$, suggesting the challenge of data sparsity in real-world.}
    \label{tab:online_strategy_comparison_full}
    \resizebox{\textwidth}{!}{
    \begin{tabular}{llclcccr}
        \toprule
        \textbf{Subject} & \textbf{Empirical Best} & \textbf{Mean Score} & \textbf{Model Predicted} & \textbf{Pred. Score} & \textbf{Sample} & \textbf{Conf.} & \textbf{Gap} \\
        \midrule
        Math       & Feynman         & 0.543 & Coach           & 0.535 & 12 & 0.082 & \textbf{0.008} \\
        French          & Coach           & 0.656 & Coach           & 0.656 & 6  & 0.086 & \textbf{0.000} \\
        Geography & Coach           & 0.737 & Coach           & 0.737 & 5  & 0.076 & \textbf{0.000} \\
        Biology       & Coach           & 0.628 & Coach           & 0.628 & 10 & 0.089 & \textbf{0.000} \\
        PhysChem           & Feynman         & 0.644 & Baseline        & 0.455 & 16 & 0.085 & 0.189 \\
        History   & Baseline        & 0.612 & Coach           & NaN   & 4  & 0.079 & NaN \\
        Dutch     & Baseline        & 0.638 & Coach           & NaN   & 3  & 0.074 & NaN \\
        Physics    & Baseline        & 0.458 & Coach           & NaN   & 2  & 0.076 & NaN \\
        English         & Baseline        & 0.559 & Coach           & NaN   & 2  & 0.077 & NaN \\
        Economy       & Feynman         & 0.697 & Coach           & NaN   & 1  & 0.080 & NaN \\
        \midrule
        \multicolumn{8}{l}{\textbf{Overall Strategy Alignment Rate (for $n \ge 5$): 80.0\%}} \\
        \bottomrule
    \end{tabular}
    }
\end{table*}

\begin{table}[H]
    \centering
    \caption{Statistical comparison of strategies. \textit{Baseline} is the production prompt; \textit{Feynman} was favored in synthetic simulation; \textit{Coach} emerged as the empirical optimum after one month of live deployment, confirming successful sim-to-real adaptation.}
    \label{tab:strategy_pvalues}
    \begin{small}
    \begin{tabular}{lcccc}
        \toprule
        \textbf{Strategy Name} & \textbf{Mean Reward} & \textbf{Std. Dev.} & \textbf{Sample Size ($n$)} \\
        \midrule
        Baseline (Socratic) & 0.568 & 0.190 & 389 \\
        Coach (Scaffolding) & \textbf{0.622} & \textbf{0.174} & 45 \\
        Feynman (Explanation) & 0.558 & 0.185 & 164 \\
        \midrule
        \multicolumn{4}{l}{\textbf{Statistical Significance (Mann-Whitney U Test)}} \\
        \midrule
        Coach vs. Feynman & \multicolumn{3}{l}{$p = 0.0253$ (\textbf{Significant})} \\
        Coach vs. Baseline & \multicolumn{3}{l}{$p = 0.0675$ (Marginally Significant)} \\
        Feynman vs. Baseline & \multicolumn{3}{l}{$p = 0.4489$ (Not Significant)} \\
        \bottomrule
    \end{tabular}
    \end{small}
\end{table}
\newpage

\section{Comparison of Conversion Rate in Real-World Deployment}
\label{app:comp_conversion_real}

\begin{figure}[H]
    \centering
    \begin{subfigure}[b]{0.48\textwidth}
        \centering
        \includegraphics[width=\linewidth]{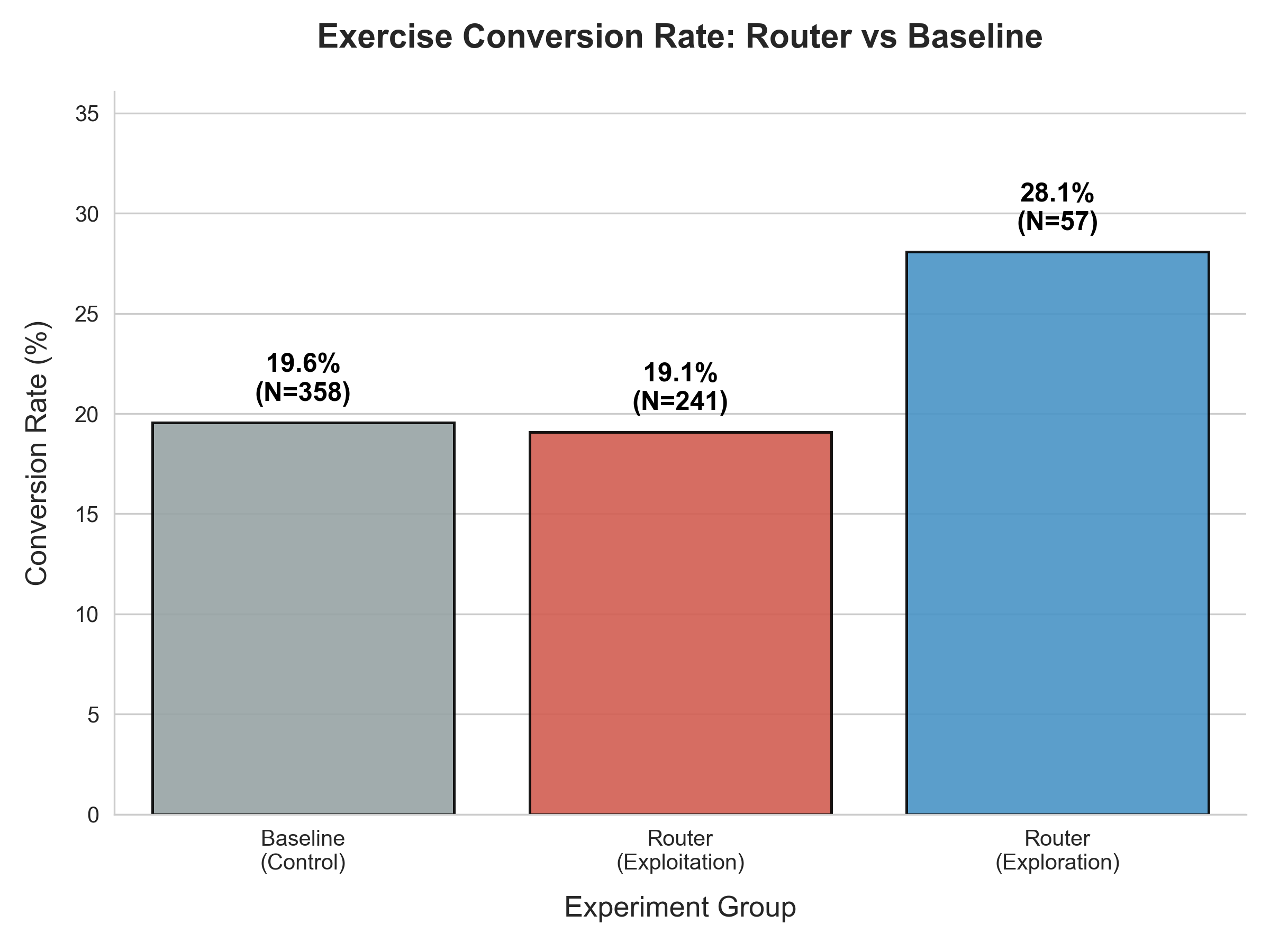}
        \caption{Detailed Experiment Groups}
        \label{fig:conversion_groups_detailed}
    \end{subfigure}
    \hfill 
    \begin{subfigure}[b]{0.48\textwidth}
        \centering
        \includegraphics[width=\linewidth]{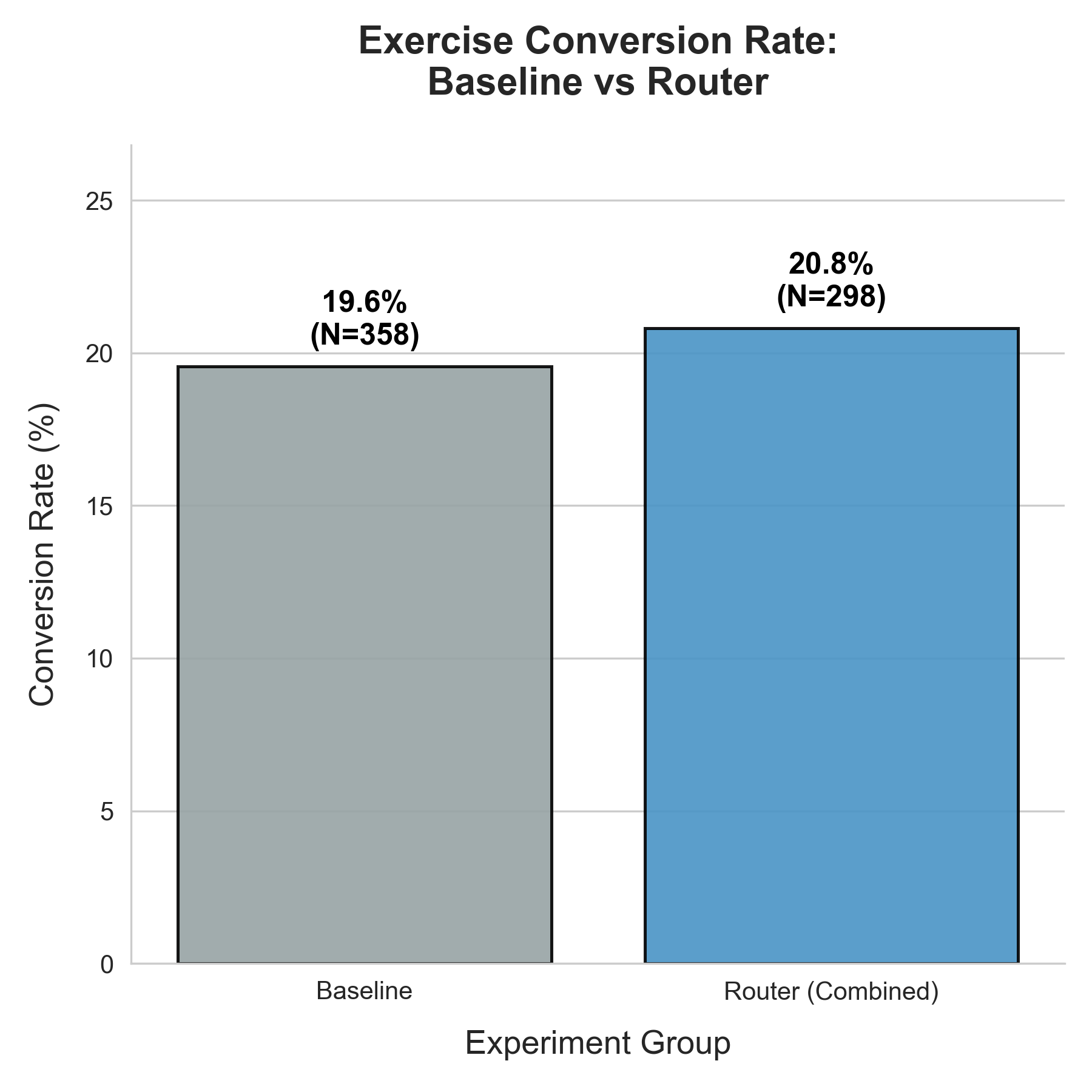}
        \caption{Baseline vs. Router (Combined)}
        \label{fig:conversion_groups_combined}
    \end{subfigure}
    
    \caption{Comparison of Exercise Conversion Rates. (a) The detailed breakdown reveals that the exploration group achieved a promising conversion rate (28.1\%), indicating the router's potential to discover engaging strategies. (b) When aggregating the router groups (exploitation and exploration), the dynamic routing system yields an overall conversion rate of 20.8\%, achieving a 1.2\% absolute improvement compared to the static baseline (19.6\%).}
    \label{fig:conversion_comparison}
\end{figure}

Comparing the router with the baseline as Figure~\ref{fig:conversion_comparison}, we observe that while the exploitation group ($19.1\%$) performs similarly to the baseline ($19.6\%$), the exploration group ($28.1\%$) outperforms the others. By combining router groups, the overall conversion rate of the router is $20.8\%$, yielding a $1.2\%$ absolute improvement compared to the baseline. However, the system-level improvement did not reach statistical significance via the Chi-square test ($p=0.76$). This result might be because of the constrained sample size during the live deployment; the promising results from the exploration phase demonstrate the routing model's potential to discover and adapt to highly engaging pedagogical strategies over time.

\newpage
\section{Comparison of Conversion Rate in Real-World Deployment}
\label{app:comp_accuracy}
\begin{figure}[H]
    \centering
    \includegraphics[width=0.5\linewidth]{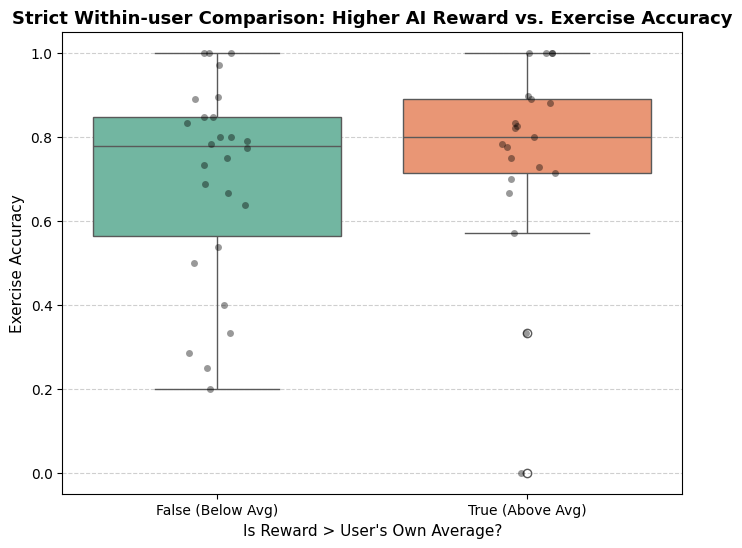}
    \caption{Distribution of student exercise accuracy categorized by AI score brackets. Although the difference did not reach statistical significance ($p = 0.3736$), sessions following high-score interactions show a positive performance trend ($0.760$ vs. $0.700$). The concentration of higher scores in the high-score group suggests that while the impact on immediate accuracy is gradual, the system provides a more consistent learning foundation for students.}
    \label{fig:comparison_score_score}
\end{figure}

\begin{figure}[H]
    \centering
    \includegraphics[width=0.6\linewidth]{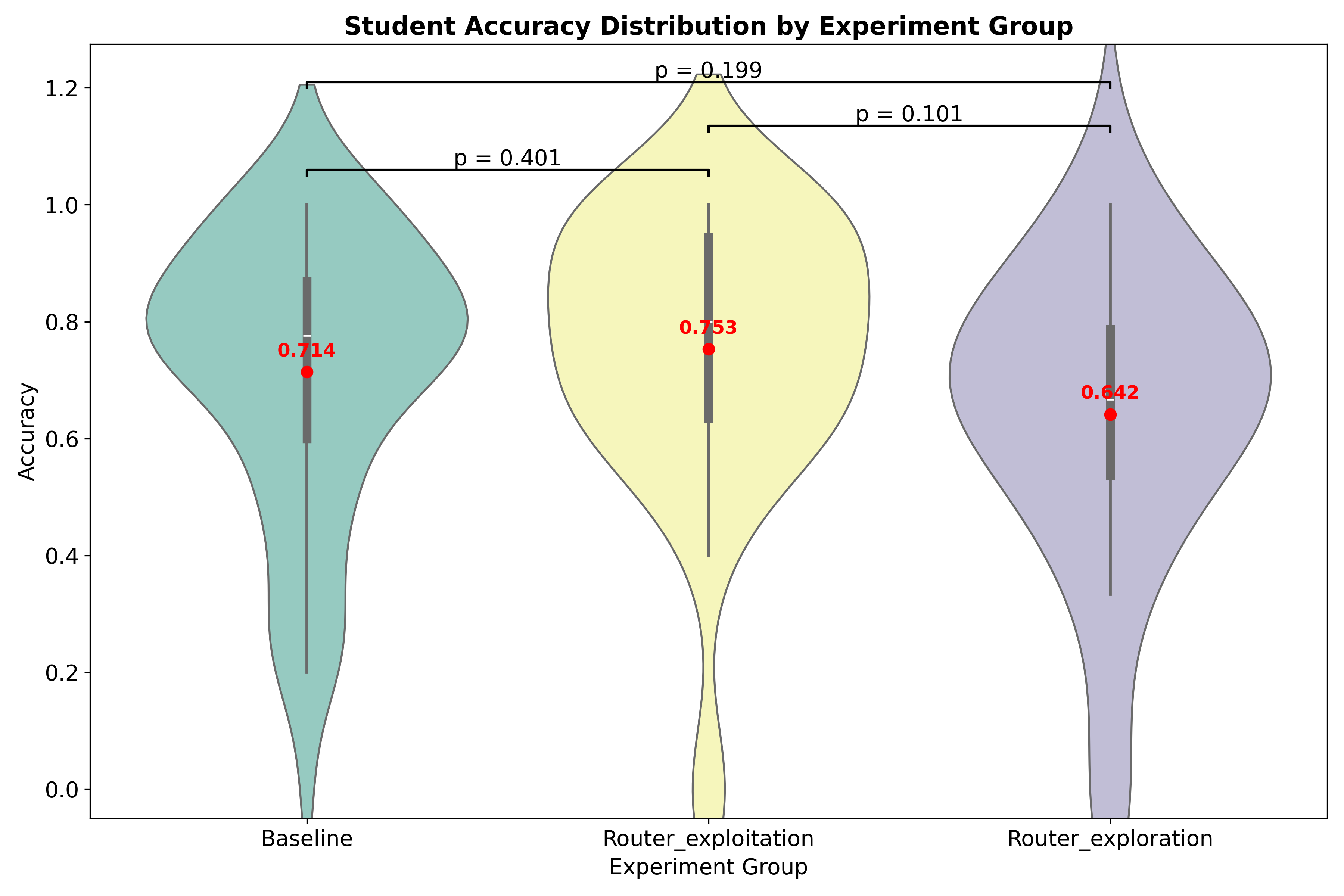}
    \caption{Comparison of Exercise Accuracy. The exploitation group of router shows the highest average accuracy $75.3\%$, but the reason could be students' inherent ability. Hence, we tend to apply within-user analysis.}
    \label{fig:comp_accuracy}
\end{figure}

Finally, we examined whether high-quality interactions translate to higher accuracy (Figure \ref{fig:comparison_score_score}). While a positive trend was observed, with high-score sessions yielding a $6\%$ higher mean accuracy ($0.760$ vs. $0.700$), the result was not statistically significant ($p = 0.3736$), possibly due to the small sample size $N=56$.
Moreover, this suggests a potential ``scaffolding paradox'': while the AI successfully guides the student through complex concepts (as reflected in the high conversion), the knowledge gain is a longitudinal process that may not be captured in a single-session snapshot. This finding underscores the role of the AI tutor as an engagement catalyst rather than a mere answer generator.
Furthermore, we compared the exercise accuracy across three groups in the live deployment (Figure~\ref{fig:comp_accuracy}). While the exploitation achieves the highest average accuracy $75.3\%$, compared to baseline and exploration ($71.4\%$ and $64.3\%$), the p-value is not significant. However, the difference might be due to the students' inherent level, which is why we applied the within-user analysis. Unfortunately, it further reduces the sample size since the data only allowed students with multiple exercises, illustrating the real-world challenge of data sparsity.

\newpage
\section{Simulated Conversation Transcripts}
\label{app:sim_conversation} 
To provide qualitative insights into the dynamics of our simulated environment, we present representative examples of generated conversations. Original interactions (in Dutch) have been translated into English without manual polishing to authentically reflect the raw outputs of the training environment. 

\begin{tcolorbox}[
    colback=gray!5!white,        
    colframe=gray!50!black,     
    title=\textbf{Example 1: Strategy Adaptation with an Unmotivated Student} 
]
\textbf{Metadata:}
\begin{itemize}
    \setlength{\itemsep}{1pt}
    \item \textit{Subject:} French (Personal Pronouns: \textit{lui/leur})
    \item \textit{Student Profile:} Low Motivation
    \item \textit{Selected Prompt:} Analytical
    \item \textit{Final Score:} 0.425
\end{itemize}
\vspace{-0.5em}
\rule{\linewidth}{0.4pt} 
\vspace{0.5em}

\textbf{Tutor:} The topic is personal pronouns as indirect objects. It is about how you replace people after a verb with `à` with `lui` or `leur`. Can you explain in your own words when to use `lui` and when `leur`?

\vspace{0.8em}
\textbf{Student:} lui = 1 person, leur = more than one, done, lol

\vspace{0.8em}
\textbf{Tutor:} Well done! Personal pronouns as indirect object: lui = one person, leur = more than one. Remember: the pronoun always goes before the conjugated verb or the infinitive. \textit{[Internal Trigger: push\_practice\_button]} Can you explain why it's important that the pronoun stays there?

\vspace{0.8em}
\textbf{Student:} huh, don't care — otherwise lui/leur gets lost in the sentence lol

\vspace{0.8em}
\textbf{Tutor:} \textit{[Uses Analogy]} In short — think of lui/leur as an anchor that pulls the person forward so they don't get ``lost'' at the back of the sentence. In the sentence ``Solange a écrit à ses amies,'' which pronoun do you use and why?

\vspace{0.8em}
\textbf{Student:} don't care, I want pizza 

\vspace{0.8em}
\textbf{Tutor:} \textit{[Humorous Redirection]} It's great you're thinking about pizza! We use this with pizza too. Usually, lui/leur is before the verb, but in a command (imperative), it comes after the verb with a hyphen. In the sentence ``Donne la pizza à Marc!'', which pronoun do you use and where do you put it?

\vspace{0.8em}
\textbf{Student:} lui, after the verb: Donne-lui la pizza! oh really?? lol ok
\end{tcolorbox}
\newpage
\begin{tcolorbox}[
    colback=gray!5!white,
    colframe=gray!50!black,
    title=\textbf{Example 2: Sub-optimal Strategy Selection (Strategy Mismatch)}
]
\textbf{Metadata:}
\begin{itemize}
    \setlength{\itemsep}{1pt}
    \item \textit{Subject:} Economics (Where does money come from?)
    \item \textit{Student Profile:} Low Motivation
    \item \textit{Selected Prompt:} Baseline\_opt
    \item \textit{Final Score:} 0.268
\end{itemize}
\vspace{-0.5em}
\rule{\linewidth}{0.4pt}
\vspace{0.5em}

\textbf{Tutor:} What do you think the topic "Where does money come from?" investigates?

\vspace{0.8em}
\textbf{Student:} huh what? don't care, think banks or leprechauns printing money lol

\vspace{0.8em}
\textbf{Tutor:} Haha, leprechauns are a great explanation in fairy tales — but for "Where does money come from?" we look at real processes. Do you think banks really just "print" money when they give loans, or is something else going on? Why?

\vspace{0.8em}
\textbf{Student:} yeah kind of, banks create scriptural money (girale tegoeden) when they give loans and thus create money, not really banknotes tho lol

\vspace{0.8em}
\textbf{Tutor:} Well done. If commercial banks create scriptural credit when providing loans, why do you think central banks like the ECB and DNB need to control this process? What are they trying to achieve?

\vspace{0.8em}
\textbf{Student:} so they can grab more money for themselves lol
\end{tcolorbox}

\begin{tcolorbox}[
    colback=gray!5!white,
    colframe=gray!50!black,
    title=\textbf{Example 3: Sustained Scaffolding with a Moderate Student}
]
\textbf{Metadata:}
\begin{itemize}
    \setlength{\itemsep}{1pt}
    \item \textit{Subject:} Mathematics (Calculating growth factors)
    \item \textit{Student Profile:} Moderate
    \item \textit{Selected Prompt:} Analytical\_opt
    \item \textit{Final Score:} 0.794
\end{itemize}
\vspace{-0.5em}
\rule{\linewidth}{0.4pt}
\vspace{0.5em}

\textbf{Tutor:} Good question — this topic is "Calculating growth factors for different time units". A quick recap: we use $N = b \cdot g^t$ where $g$ is the growth factor per given time unit. To get the growth factor for a different duration, we raise $g$ to the ratio of those times. How would you determine the growth factor per 30 minutes if $g$ is given per hour — and why exactly must you use that power operation?

\vspace{0.8em}
\textbf{Student:} Ahhh ok just $g^{(1/2)} = \sqrt{g}$, because 30 min is 0.5 hour. Because growth per time unit multiplies, so half time = taking the root lol.

\vspace{0.8em}
\textbf{Tutor:} Well done — exactly: $g^{(1/2)} = \sqrt{g}$ because 30 minutes equals 0.5 hours. Your logic is sound. \textit{[Internal Trigger: push\_practice\_button]} If $g = 1.2$ per hour, which growth factor belongs to 15 minutes and why is $g^{(1/4)}$ the right choice instead of simply $g/4$?

\vspace{0.8em}
\textbf{Student:} Ahhh ok, for 15 min it's $g^{(1/4)}=1.2^{(1/4)} \approx 1.047$ because 15 min = 1/4 hour and growth works multiplicatively so you take the 4th root, not $g/4$ because that would be linear division lol.

\vspace{0.8em}
\textbf{Tutor:} Well done — your logic is sound. If you now know the growth factor for 15 minutes (e.g., 1.047), how do you algebraically calculate back to the growth factor per hour, and which power rule explains that these two are consistent?

\vspace{0.8em}
\textbf{Student:} If the growth factor per 15 min is $g15$ then the growth factor per hour is $g = g15^4$, because $(a^m)^n = a^{mn}$ so $(g^{(1/4)})^4 = g$ — easy lol.

\vspace{0.8em}
\textbf{Tutor:} Well done! Can you now pick a concrete number (e.g., $g15 = 1.047$), calculate what $g$ per hour becomes, and explain how those calculations clarify the meaning of the power rule? \textit{[Internal Trigger: push\_practice\_button]}

\vspace{0.8em}
\textbf{Student:} No idea lol, don't get this at all, can you explain it simply?

\vspace{0.8em}
\textbf{Tutor:} \textit{[Scaffolding Fallback]} Good that you mention it, let's keep it simple. Think of 1 hour as four pieces of 15 minutes: if each piece multiplies by $x$ and after four pieces you have your total growth factor per hour $g$, then $x^4 = g$, so $x = g^{(1/4)}$ (the fourth root). Now explain in your own words why we use exponentiation with a fraction instead of simply dividing $g$ by 4?

\vspace{0.8em}
\textbf{Student:} Because growth multiplies, it doesn't add up — you look for $x$ with $x^4 = g$ so $x = g^{(1/4)}$, and $g/4$ would cut it linearly and after four pieces wouldn't lead to $g$, but something completely different lol.
\end{tcolorbox}
\newpage
\section{Qualitative Analysis: Real-World Scaffolding}
\label{app:real_chat}
To examine the router's decision-making logic, we analyze a real-world interaction (Geography; Score: 0.82) below. Although the router selects the system prompt only once at the session onset, this case highlights that the adaptive prompt identifies resilient pedagogical templates. The interaction shows that the model-selected \textit{Coach} prompt enables the LLM to maintain a consistent instructional trajectory despite real-world noise, such as interpreting student typos without breaking flow. Most importantly, it exhibits autonomous level adaptation: when the student repeatedly struggles with the abstract ``uniformitarianism principle,'' the LLM, guided by the model-selected stance, turns to a concrete volcano analogy to reduce cognitive load. 

\begin{tcolorbox}[
    colback=gray!5!white,
    colframe=gray!50!black,
    title=\textbf{Case Study: Real-World Scaffolding in Geography}
]
\textbf{Metadata:}
\begin{itemize}
    \setlength{\itemsep}{1pt}
    \item \textit{Subject:} Geography (Earth's History)
    \item \textit{Environment:} Real-world Deployment
    \item \textit{Selected Prompt:} Coach
    \item \textit{Final Score:} 0.82
\end{itemize}
\vspace{-0.5em}
{\scriptsize \textit{(Note: Translated from Dutch; user inputs slightly paraphrased for privacy while preserving original errors and tone.)}}
\vspace{0.2em}
\rule{\linewidth}{0.4pt}
\vspace{0.5em}

\textbf{Tutor:} [...] Which rock is formed by the compression of sediment layers over a long time?

\vspace{0.8em}
\textbf{Student:} Argumentation rock [Typo: meant Sedimentary]

\vspace{0.8em}
\textbf{Tutor:} \textit{[Correction \& Encouragement]} Good effort! You probably mean "sedimentary rock." [...] Why are fossils often found in these layers?

\vspace{0.8em}
\textbf{Student:} yeah sorry autocorrect. igneous rock melts things, but sedimentary is soft so things stay intact better.

\vspace{0.8em}
\textbf{Tutor:} \textit{[Scaffolding]} Well explained! Now, can you explain what the uniformitarianism principle (actualiteitsprincipe) means?

\vspace{0.8em}
\textbf{Student:} I don't know that.

\vspace{0.8em}
\textbf{Tutor:} \textit{[Level Adaptation]} It's okay! It means processes we see today (erosion, volcanoes) also happened in the past. Why is it handy to use current processes to study old rocks?

\vspace{0.8em}
\textbf{Student:} I still don't get it.

\vspace{0.8em}
\textbf{Tutor:} \textit{[Concrete Analogy]} Imagine a volcano spewing lava today; it cools into rock. We know it worked the same way in the past. Does this mean the Earth worked very differently then, or similarly to now?

\vspace{0.8em}
\textbf{Student:} Very similarly to now.

\end{tcolorbox}

\end{document}